%% file: main.tex
\documentclass[10pt,twocolumn,letterpaper]{article}
\usepackage[utf8]{inputenc}
\usepackage[T1]{fontenc}

\providecommand{\BuildMode}{1}
\newif\ifbuildmain
\newif\ifbuildsupp
\buildmainfalse
\buildsuppfalse
\ifnum\BuildMode=0\relax
  \buildmaintrue
\fi
\ifnum\BuildMode=1\relax
  \buildmaintrue
  \buildsupptrue
\fi
\ifnum\BuildMode=2\relax
  \buildsupptrue
\fi

\usepackage[algorithms]{wacv}

\usepackage{bm,mathtools}
\usepackage{multirow}
\usepackage{wrapfig}
\usepackage{makecell}
\usepackage{threeparttable}
\usepackage{siunitx}
\usepackage{array}
\usepackage{longtable}
\usepackage{tabularx}
\usepackage{placeins}  %
\usepackage{fancyvrb}  %
\usepackage{needspace}  %

\usepackage[accsupp]{axessibility}  %

\providecommand{\linelabel}[1]{}  %
\newif\ifrevblue \revbluefalse
\newcommand{\revblue}[1]{\ifrevblue{\color{blue}#1}\else#1\fi}

\definecolor{wacvblue}{rgb}{0.21,0.49,0.74}
\usepackage[pagebackref,breaklinks,colorlinks,allcolors=wacvblue]{hyperref}

\usepackage{xr-hyper}
\edef\ThisJob{\jobname}
\def\MainJob{main}
\def\SuppJob{supp}
\ifnum\BuildMode=0\relax
  \ifx\ThisJob\SuppJob\else
    \IfFileExists{supp.aux}{\externaldocument[][nocite]{supp}}{}
  \fi
\fi
\ifnum\BuildMode=2\relax
  \ifx\ThisJob\MainJob\else
    \IfFileExists{main.aux}{\externaldocument[][nocite]{main}}{}
  \fi
\fi

\ifnum\BuildMode=1\relax\else
  \makeatletter
  \def\NoFileLink#1#2#3{#1\Hy@xspace@end}
  \AtBeginDocument{\let\hyper@linkfile\NoFileLink}
  \makeatother
\fi
\def\wacvPaperID{1641}
\def\confName{WACV}
\def\confYear{2027}

\title{LoC-Path: Learning to Compress for Pathology Multimodal Large Language Models}

\author{Qingqiao Hu$^{1}$, Weimin Lyu$^{2}$, Meilong Xu$^{1}$, Xiaoling Hu$^{3}$, Kehan Qi$^{1}$, Saumya\\ Gupta$^{1}$, Jiawei Zhou$^{1}$, and Chao Chen$^{1}$\\[0.6em]
$^{1}$Stony Brook University \quad $^{2}$Amazon\\
$^{3}$Athinoula A. Martinos Center for Biomedical Imaging,\\ Massachusetts General Hospital and Harvard Medical School\\
{\tt\small qingqiao.hu@stonybrook.edu}
}

\definecolor{BaselineBlue}{RGB}{33,102,172}
\definecolor{OursRed}{RGB}{180,15,32}
\newcommand{\myparagraph}[1]{\smallskip\noindent\textbf{#1}}

\newcommand{\suppsec}[1]{Supp.\ Sec.~\ref{#1}}
\newcommand{\supptab}[1]{Supp.\ Tab.~\ref{#1}}

\begin{document}

\ifbuildmain
   \maketitle
\begin{abstract}
Whole Slide Image (WSI) multimodal large language models (MLLMs) are difficult to build and deploy because gigapixel slides induce thousands of \revblue{highly redundant} visual tokens \revblue{that must be learned from sparse slide-level language supervision}. Existing slide-level pathology MLLMs typically combine heavy slide-level encoders with long visual prefixes, making development and deployment expensive under limited computational resources. We revisit this regime and show that WSI tile features are highly redundant at both global and local scales, while task-relevant evidence is sparse and query-dependent. We therefore introduce \textbf{LoC-Path}, a resource-efficient slide-level MLLM that compresses before fusion. LoC-Path uses a sparse token merger and a masked-autoencoder (MAE)-pretrained resampler to replace expensive slide-level encoding with a compact latent interface. A token importance scorer selects the most relevant latents and cross-attention routing adapters fuse them into a few LLM decoder layers. This design lowers both multimodal tuning cost and inference-time latency/memory by avoiding heavy slide-level encoding and long visual prefixes. Extensive experiments show that the inference TFLOPs of LoC-Path \revblue{are} \textbf{\revblue{97}\%} lower and inference memory is also \textbf{39\%} lower. LoC-Path enables \revblue{efficient} end-to-end development and deployment under limited computational resources while remaining competitive with prior slide-level MLLMs. Code is available at \mbox{\href{https://github.com/WinstonHuTiger/loc_path}{\texttt{github.com/WinstonHuTiger/loc\_path}}}.
\end{abstract}

\vspace{-.15in}
\section{Introduction}
\label{sec:intro}
Digital pathology has undergone a significant revolution in recent years. Whole Slide Images (WSIs), with many exceeding $10,000\times10,000$ pixels, are partitioned into small patches.\footnote{Throughout this paper, we use \emph{tile} and \emph{patch} interchangeably.} Powerful pretrained image encoders~\cite{lu2024avisionlanguage,chen2024towards,vorontsov2024foundation} are applied to these patches to extract patch-wise features. These features are then aggregated through an attention mechanism to make a whole-slide level prediction~\cite{lu2021data,li2021dual,zhang2024attention}, whereas the attention will highlight a potentially small set of critical patches, imitating how pathologists make clinical decisions. 

The advent of large language models (LLMs) has expanded WSI analysis beyond classification to richer outputs such as visual question answering and pathology report generation. However, whole-slide MLLMs operate in a regime that differs fundamentally from standard vision-language settings: the visual input sequence is extremely long (often thousands of tile tokens), slide-level text supervision is weak (many tiles map to one report), and diagnostically relevant evidence is sparse. Most existing WSI-level pathology MLLMs follow a LLaVA-style late-fusion design~\cite{liu2023llava} \revblue{(``late'' in the MLLM sense: projected visual tokens are concatenated with the text tokens and processed jointly by the LLM's self-attention)}: they convert patch features into a long visual token sequence, concatenate it with text tokens, and perform self-attention over the joint sequence inside the LLM. This strategy is adopted by multiple slide-level MLLMs, including WSI-LLaVA and SlideChat~\cite{liang2025wsi,chen2025slidechat}. Other variants, such as ALPaCA~\cite{gao2025alpaca} and TCP-LLaVA~\cite{lyu2025efficient}, reduce the number of visual tokens entering the LLM with a learned bottleneck (e.g., BLIP-2's Q-Former~\cite{li2023blip2}). Nevertheless, they still operate under the same WSI regime of extreme sequence length, weak slide-text alignment, and sparse task-relevant evidence. Thus, the LLaVA-style concatenate-and-attend strategy becomes increasingly inefficient and difficult to train under weak slide-level supervision.

\begin{figure*}[t]
  \centering
  \begin{subfigure}[t]{0.24\textwidth}
    \centering
    \includegraphics[width=\linewidth]{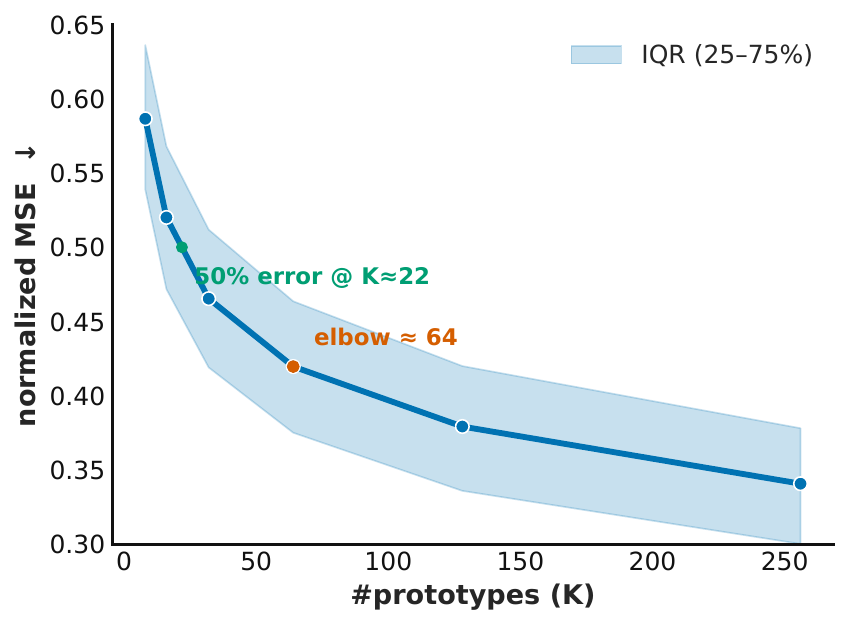}
    \caption{Compression.}
    \label{fig:kcurve}
  \end{subfigure}\hfill
  \begin{subfigure}[t]{0.24\textwidth}
    \centering
    \includegraphics[width=\linewidth]{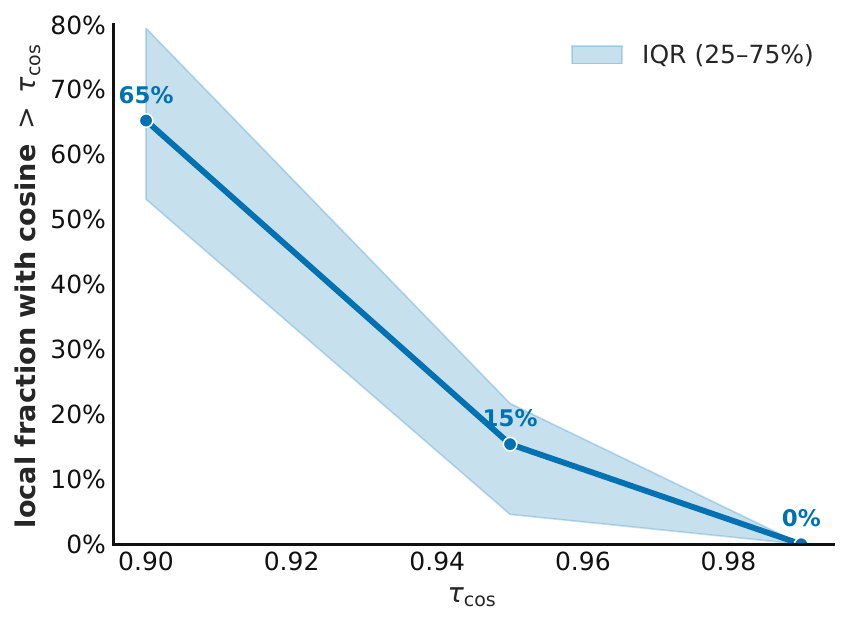}
    \caption{Local redundancy.}
    \label{fig:redundancy}
  \end{subfigure}\hfill
  \begin{subfigure}[t]{0.24\textwidth}
    \centering
    \includegraphics[width=\linewidth]{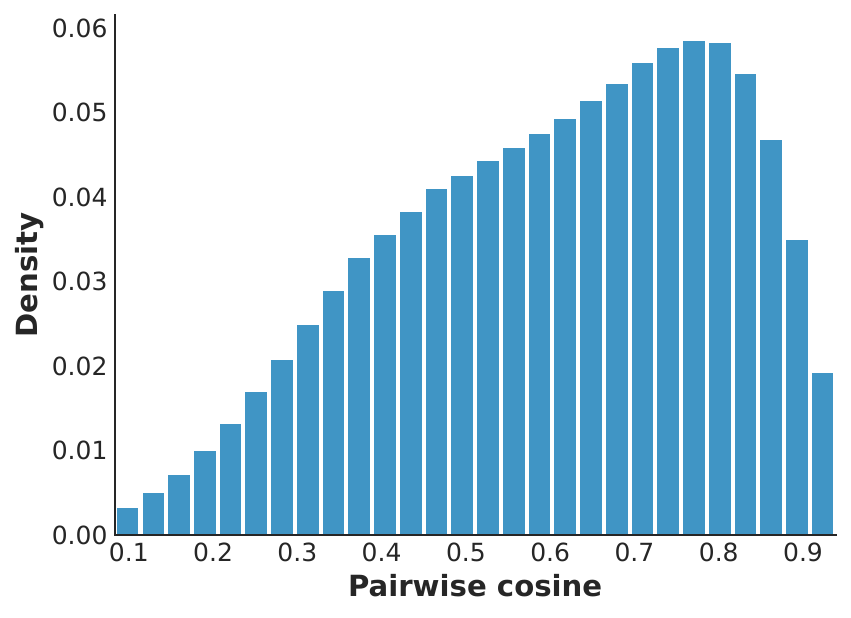}
    \caption{Pairwise similarity.}
    \label{fig:paircos}
  \end{subfigure}\hfill
  \begin{subfigure}[t]{0.24\textwidth}
    \centering
    \includegraphics[width=0.86\linewidth]{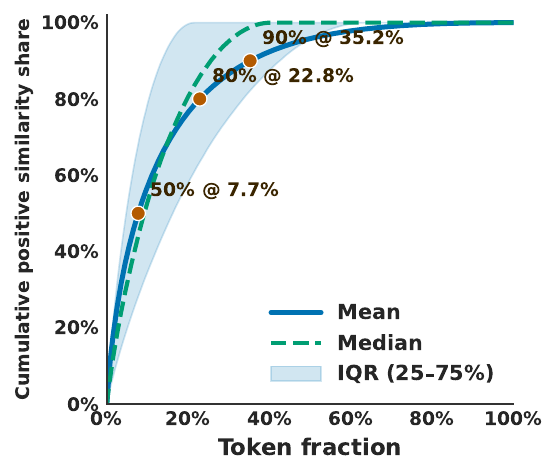}
    \caption{Task relevance.}
    \label{fig:cap_align_redundancy}
  \end{subfigure}
  \vspace{-1em}
  \caption{\textbf{WSI tile tokens are redundant globally and locally, and only a sparse subset is task-relevant.}
  We analyze $\sim$2k TCGA WSIs using CONCH-V1~\cite{lu2024avisionlanguage}. (a) A small prototype budget reconstructs tile features well. (b,c) Local-neighbor and pairwise statistics both show substantial redundancy. (d) A small token subset carries most report-aligned signal: $7.7\%$, $22.8\%$, and $35.2\%$ of tokens explain $50\%$, $80\%$, and $90\%$ of the positive mass.}
  \label{fig:wsi_redundancy_suite}
  \vspace{-1.5em}
\end{figure*}

This mismatch creates not only an algorithmic bottleneck but also a practical resource bottleneck. On the vision side, many methods rely on additional slide-level aggregation over long tile sequences to recover global context. Slide-level encoders are usually expensive in memory and \revblue{floating-point operations (}FLOPs\revblue{)}, making end-to-end slide-level MLLM development difficult under limited computational resources. On the language side, concatenating long visual sequences with text within the LLM, as LLaVA-style fusion~\cite{liu2023llava}, further increases cost during multimodal training and inference, even if the LLM backbone is frozen.

\myparagraph{Preliminary analysis: redundancy and task irrelevance in visual tokens.}
To fully understand the underlying issues, we carried out a preliminary quantitative analysis, which shows that most WSI tokens are redundant, and only a sparse subset is task-relevant in Fig.~\ref{fig:wsi_redundancy_suite}.

\myparagraph{Issue 1: redundancy.} We observe that the tile-level token sequences of WSIs are highly redundant both \emph{globally} and \emph{locally}. We analyze the tile features of the $2{,}000$ TCGA~\cite{weinstein2013cancer} WSIs and show in Fig.~\ref{fig:wsi_redundancy_suite}: 
(1)~a small set of $K\!\approx\!250$ prototype groups can fairly well reconstruct the entire tile feature set; 
and (2)~tile features are highly similar to each other, as observed by comparing neighboring tiles and comparing pairs of tiles far apart.  
These findings confirm strong redundancy in both global content and local neighborhoods, suggesting that WSIs can be effectively represented by compact latent representations rather than full sequences of thousands of tokens.

\myparagraph{Issue 2: task irrelevance.}
We also observe that only a limited number of tiles are truly relevant to the task, i.e., relevant to the text supervision (answers in the VQA task, and report content in the report generation task). We use the same $2{,}000$ TCGA WSIs and compute the cosine similarity between each tile token and the corresponding text supervision embedding. We rank tokens by their descending cosine similarity with the text supervision and compute the cumulative sum of positive similarity as a function of total token fraction.
Shown in Fig.~\ref{fig:cap_align_redundancy}, the top $35.2\%$ of tile tokens account for $90\%$ of the total cosine similarity. In other words, a majority of visual tokens are irrelevant to the pathology report. This aligns with how pathologists rely on key areas of WSIs for diagnosis.
Details are in \suppsec{app:redundancy}.

\myparagraph{Our contributions.}
Inspired by these observations, we target slide-level MLLM efficiency from the perspective of both end-to-end \emph{development} and \emph{deployment}.\footnote{\revblue{Following standard practice in WSI modeling, tile-level features are extracted once by a frozen CONCH encoder as preprocessing;} ``end-to-end'' refers to the unified \emph{slide-level} stack from tile embeddings to text outputs\revblue{, which is trained jointly and used unchanged at inference}.}\linelabel{rl:R01-scope} We name our efficient framework \textbf{LoC-Path} that combines multi-stage compression before fusion with routed cross-attention during fusion. Window-based Sparse Token Merger (\textbf{STM}) and the MAE-pretrained resampler reduce the slide representation from thousands of tile tokens to a compact latent set, lowering the cost of building and tuning slide-level models. Token Importance Scorer (\textbf{TIS}) and Cross-Attention Routing Adapter (\textbf{CARA}) modules further limit the LLM-side cost by selecting only a small set of query-relevant latents into a few decoder layers, instead of concatenating long visual prefixes.

As mentioned, existing LLaVA-style fusion methods use self-attention on concatenated visual and text tokens~\cite{liu2023llava}. We stipulate that this is unnecessary, and propose to use cross-attention across visual and text tokens instead. Let $T$ denote the text prefix length and $L$ the number of visual tokens.
LLaVA-style fusion applies self-attention over the concatenated sequence, incurring $\mathcal{O}\!\left((T+L)^2\right)$ attention cost per layer.
In contrast, our design keeps self-attention on text only and injects vision via cross-attention on a routed set of $M\le L$ latents, with cost $\mathcal{O}(T^2 + T M)$ (see \suppsec{sup:theo_complexity_cara} for more theoretical analysis).
We inject TIS-selected visual tokens into multiple decoder layers of the LLM through CARA modules equipped with learnable gates. Multi-layer routing allows richer supervision from multiple depths of the LLM, improving alignment while reducing computation. 

In summary, our main contributions are as follows.
\begin{itemize}
\item \textbf{WSI regime characterization.}
We identify a distinct whole-slide regime for MLLMs: strong multi-scale redundancy within a slide (Fig.~\ref{fig:wsi_redundancy_suite}) and sparse, query-dependent task relevance (Fig.~\ref{fig:cap_align_redundancy}), which breaks the standard assumption that most visual tokens are worth preserving and jointly attending with text.
\item \textbf{A resource-efficient slide-level MLLM.}
    We propose \textbf{LoC-Path}, which combines local token merging, global latent compression, and query-dependent routed fusion. 
\item \textbf{End-to-end efficiency with on-par quality.} These two careful design choices enable end-to-end efficiency for both development and deployment while matching the strongest slide-level MLLM on average at a fraction of the compute.
\end{itemize}

\begin{figure*}[t]
    \centering
	    \includegraphics[width=0.95\textwidth]{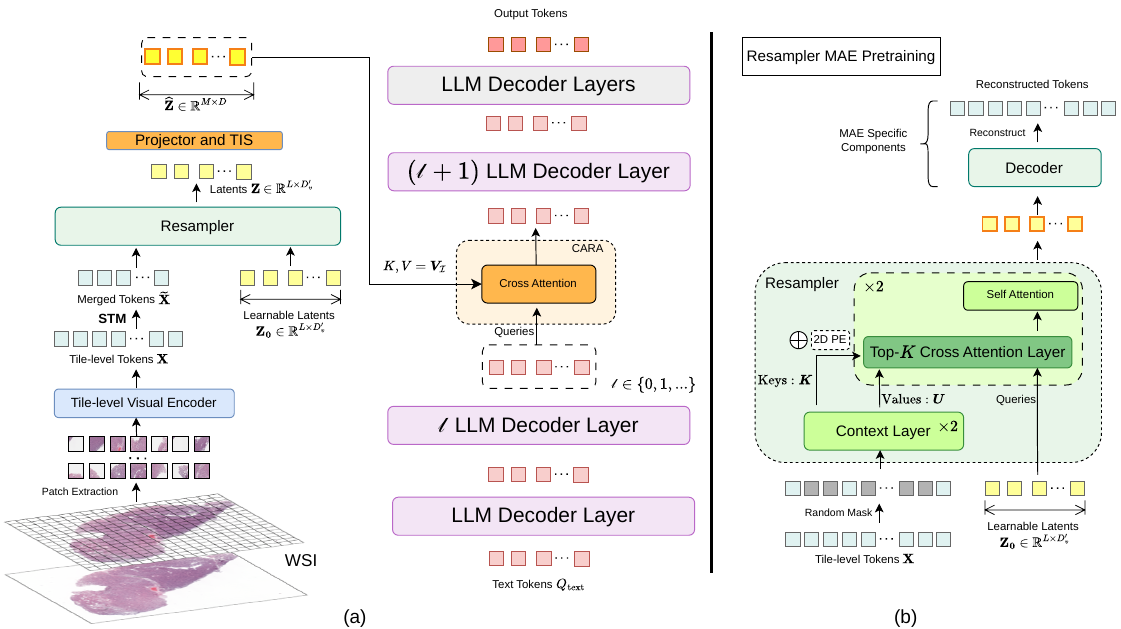}
	    \vspace{-1em}
	        \caption{
	        Overview of \textbf{LoC-Path}. \textbf{(a)} End-to-end pipeline with STM+Resampler and routed fusion (TIS/CARA). STM and resampler are designed to reduce WSI's tile-level tokens redundancy locally and globally; TIS and CARA modules are designed to fuse the visual tokens with LLM in a query-dependent manner through cross-attention. \textbf{(b)} Resampler's MAE pretraining helps it learn to glance through the tile-level token sequence and then collectively compress it. See Sec.~\ref{sec:mae_resampler} and Sec.~\ref{sec:fusion} for details.}

	    \label{fig:main_frame}
	    \vspace{-1.5em}
	\end{figure*}

\vspace{-.5em}\section{Related Work}

\myparagraph{Pathology Foundation Models.}
Pathology foundation models form two categories. \emph{(i) Tile-level} encoders, self-supervised on large histopathology corpora, span early contrastive models~\cite{wang2022ctranspath,huang2023plip,ikezogwo2023quilt}, larger DINOv2-based~\cite{oquab2023dinov2} foundation models~\cite{chen2024towards,vorontsov2024foundation,zimmermann2024virchow2,filiot2024phikonv2,nechaev2024hibou,dippel2024rudolfv,bioptimus2024hoptimus}, and vision--language variants~\cite{lu2024avisionlanguage,xiang2025musk}; they supply the per-tile embeddings as we mentioned. \emph{(ii)} To turn long tile sequences into a single slide vector, a second family of \emph{slide-level} encoders is then pretrained on frozen tile features via masked autoencoding~\cite{xu2024gigapath}, vision-to-language alignment~\cite{ding2025titan,shaikovski2024prism}, weak/molecular supervision~\cite{wang2024chief,vaidya2025moleculardrivenfoundationmodeloncologic}, multi-stain views~\cite{jaume2024madeleine}, or encoder-agnostic aggregation~\cite{lenz2025cobra}; contrastive learning on sampled tile tokens~\cite{lazard2023gigassl}. Crucially, every slide-level encoder is an \emph{additional} heavy model stacked on the tile encoder; we instead absorb its role into a lightweight resampler while maintaining its function.

\myparagraph{Multimodal LLMs in Computational Pathology.}
Patch-level assistants such as PathChat~\cite{lu2024multimodal}, Quilt-LLaVA~\cite{seyfioglu2024quilt}, PA-LLaVA~\cite{dai2024pallava}, and CPath-Omni~\cite{sun2025cpath} answer questions about individual tiles, alongside general medical-MLLM recipes~\cite{xie2025medtrinity,huang2025medvlsynther,huang2025medvlthinker} and fine-grained vision--language alignment~\cite{wang2022mgca}.
Slide-level systems~\cite{chen2025slidechat,liang2025wsi,gao2025alpaca,chen2024wsivqa,tran2025histogpt,alawode2026mllmhwsi,kim2025chatexaonepath,Huang_2026_CVPR} reason over whole slides but often inherit the two-encoder pipeline and quadratic prefilling costs discussed above.
CPathAgent~\cite{sun2025cpathagent} sidesteps the slide-level encoder by driving a single vision encoder through iterative navigation, yet its multi-turn loop can add inference latency.
Our method eliminates both the extra encoder and the multi-turn overhead in a single forward pass.

\myparagraph{Visual Token Compression for MLLMs.}
Existing token-reduction methods for MLLMs fall into three lines.
\emph{(i) Token merging} fuses similar tokens via bipartite matching (ToMe~\cite{bolya2023token}), pruning-aware fusion (ToFu~\cite{kim2024tofu}), or spectrum-preserving merges (PiToMe~\cite{tran2024pitome}).
\emph{(ii) Token pruning/selection} drops or selects tokens, e.g.\ low-attention pruning (FastV~\cite{chen2024image}), depth-progressive dropping (PyramidDrop~\cite{xing2025pyramiddrop}), or withdrawal after a chosen layer (VTW~\cite{lin2025vtw}); SparseVLM~\cite{zhang2025sparsevlm}, FitPrune~\cite{ye2025fit}, DivPrune~\cite{alvar2025divprune}, PACT~\cite{dhouib2025pact}, Dynamic-LLaVA~\cite{huang2025dynamicllava}, LLaVA-PruMerge~\cite{shang2025prumerge}, AIM~\cite{zhong2025aim}, DyCoke~\cite{tao2025dycoke}, and LLaVA-Meteor~\cite{li2025topdown} use text, attention, diversity, or learned scores.
\emph{(iii) Query-based compression} resamples the sequence into a few query tokens, as in Flamingo's Perceiver Resampler~\cite{alayrac2022flamingo}, BLIP-2's Q-Former~\cite{li2023blip2}, TokenPacker~\cite{li2025tokenpacker}, and Victor~\cite{wen2024victor}, echoing the masked-autoencoding view of distillation~\cite{he2022masked,bai2023dmae}.
Most operate as \emph{training-free, post-hoc} steps on natural images, on benchmarks whose token-reduction signal is debated~\cite{wen2025token,sun2025silent,kong2025beyond,liao2026vtcbench,shao2025tokensurvey}\revblue{; in pathology, FOCUS~\cite{guo2025focus} compresses tile tokens for few-shot WSI classification rather than for MLLM generation}. In contrast, our design pairs two \emph{trainable} mechanisms tailored to gigapixel WSIs: a query-based resampler that compresses tens of thousands of tile tokens at the vision side, and a Top-$M$ router that selects tokens \emph{inside} the LLM via a cross-attention adapter, jointly yielding linear-complexity vision--language fusion.

\vspace{-.5em}\section{Methodology}

\myparagraph{Task Formulation.}
We study WSI-level visual question answering (VQA) tasks where a whole-slide image is decomposed into a long sequence of tiles.
For each image--text pair, a tile encoder produces tile tokens
$\bm{X}=[\bm{x}_1,\ldots,\bm{x}_N]\in\mathbb{R}^{N\times D_v}$, where $N$ is the number of tiles and
$D_v$ is the tile-feature dimension. The text query is tokenized as $\bm{Q}_{\text{text}}$,
and the target is $\bm{Y}$ (e.g., VQA answers or report text), predicted as
$\hat{\bm{Y}}$ conditioned on $(\bm{X},\bm{Q}_{\text{text}})$.

\myparagraph{Notation.}
We associate each tile token $\bm{x}_i$ with a 2D grid coordinate $(c_i,r_i)$ and write
$\mathcal{C}=\{(c_i,r_i)\}_{i=1}^{N}$ (column/row indices with unit stride). In MAE pretraining, the resampler
produces $L$ visual latents $\bm{Z}\in\mathbb{R}^{L\times D_v'}$, where $L\ll N$ and $D_v'$ is the resampler
embedding dimension.

\myparagraph{Design Principles.}
WSIs operate in an extreme-length regime: the tile-token sequence is highly redundant, while task-relevant evidence
is sparse. LoC-Path is therefore built on two principles:
\textbf{(1) reduce token redundancy} by compressing the long tile sequence into a compact latent set, and
\textbf{(2) select the most task-relevant} latents for fusion with the pretrained LLM. These two principles also correspond to two practical goals: reducing the cost of building and tuning the slide-level stack, and keeping fusion cost bounded.

\myparagraph{Pipeline Overview.}
As shown in Fig.~\ref{fig:main_frame}, STM first merges the tile tokens $\bm{X}$ into $\tilde{\bm{X}}$.
An MAE-pretrained resampler then compresses $\tilde{\bm{X}}$ into latents
$\bm{Z}\in\mathbb{R}^{L\times D_v'}$. A projector adapts latent dimensions, and TIS selects Top-$M$ latents
$\hat{\bm{Z}}$ ($M\le L$) as the routed visual tokens. Finally, $\hat{\bm{Z}}$ are integrated into multiple
decoder layers of the pretrained LLM via CARA modules to produce $\hat{\bm{Y}}$.

\vspace{-.3em}\subsection{Multi-stage Token Redundancy Reduction}
\label{sec:mae_resampler}
To reduce the tile-level token redundancy, we adopt a multi-stage compression strategy. 
The key idea is inspired by how human pathologists examine WSIs. The resampler is trained through MAE pretraining, to compress token sequence into representation after scanning it broadly.

\myparagraph{Resampler for Global Redundancy Reduction.}
To reduce global redundancy in tile tokens, we adopt a resampler-style design widely used in MLLMs~\cite{alayrac2022flamingo,yu2025minicpmv45cookingefficient}. We interpret the resampler as two steps that mimic expert slide reading: \emph{scan} and \emph{compress}. First, lightweight context layers scan the full sequence and produce contextualized tokens. Then, a small set of learnable latent queries selectively aggregates distinct information from the scanned tokens via sparse (Top-$K$) cross-attention, preventing the latents from collapsing to a global average over highly redundant inputs. A subsequent latent self-attention further decorrelates the latents. The architecture is illustrated in Fig.~\ref{fig:main_frame}(b). Concretely, given tile-level tokens $\bm{X}\in\mathbb{R}^{N\times D_v}$, we compute
$\hat{\bm{X}}=\operatorname{ContextLayers}(\bm{X})$, where $\hat{\bm{X}}\in\mathbb{R}^{N\times D_v}$.
We initialize $L$ learnable latent queries $\bm{Z}_0\in\mathbb{R}^{L\times D_v'}$ and update them with $R$ stacked
Top-$K$ cross-attention blocks (LayerNorm + residual FFN). We inject 2D positional encoding only into the \textbf{keys}:
$\bm{K}=(\hat{\bm{X}}+\Phi)W_k,\;\; \bm{U}=\hat{\bm{X}}W_v^{\mathrm{res}}$,
where $\Phi\in\mathbb{R}^{N\times D_v}$ denotes 2D Fourier positional encoding and $d_h$ is the per-head key dimension.
For block $r=1,\ldots,R$: $\bm{S}_r = \operatorname{LN}(\bm{Z}_{r-1})\bm{K}^\top/\sqrt{d_h}$, \;
$\Delta\bm{Z}_r = \operatorname{softmax}\!\big(\operatorname{TopK}(\bm{S}_r)\big)\bm{U}$, \;
$\bm{Z}_r = \bm{Z}_{r-1}+\Delta\bm{Z}_r+\operatorname{FFN}\!\big(\operatorname{LN}(\bm{Z}_{r-1}+\Delta\bm{Z}_r)\big)$.
$\operatorname{TopK}$ keeps the largest $K_{\text{top}}$ attention logits per head and per query (others set to $-\infty$
before softmax), where $K_{\text{top}}$ is the sparse-attention budget. Finally, a latent self-attention layer is applied to encourage diversity, yielding resampled latents
$\bm{Z}\in\mathbb{R}^{L\times D_v'}$ with $L\ll N$.

\myparagraph{MAE Pre-training for Redundant Tile-level Tokens.}
To teach the resampler to act like a human expert, we adopt \revblue{MAE} pre-training. End-to-end training of the resampler with an LLM usually requires millions of image-text pairs, which is rare for WSIs (around $30k$ WSIs in the TCGA dataset~\cite{weinstein2013cancer} and weak supervision as shown in Fig.~\ref{fig:cap_align_redundancy}). As we have shown in Fig.~\ref{fig:kcurve}, the highly redundant tokens mean that the resampler can reconstruct masked tiles from visible context without overfitting to noise. Moreover, reconstruction on randomly masked indices prevents the resampler from outputting trivial averages and forces different latents to look at complementary regions. Concretely, we add a small decoder to use the latents for reconstruction, shown in Fig.~\ref{fig:main_frame}(b). The pre-training objective is $\mathcal{L}_{\mathrm{rec}}=\frac{1}{|\mathcal{I}_{\mathrm{mask}}|}\sum_{i\in\mathcal{I}_{\mathrm{mask}}}\|\hat{\bm{x}}_i-\bm{x}_i\|_2^2$, where $\mathcal{I}_{\mathrm{mask}}$ is the index set of masked tokens. To avoid degenerate average latents and to cover the slide, we add two regularization terms:
\vspace{-.1in}
\begin{align}
\textbf{Coverage:}\quad
\mathcal{L}_{\mathrm{cover}}
&=\frac{1}{N}\sum_{i=1}^{N}\max\!\big(0,\;\tau_{\mathrm{cov}} - u(i)\big), \label{eq:cover}\\
u(i) &= \max_z p_z(i),\\
\textbf{Diversity:}\quad
\mathcal{L}_{\mathrm{feat}}
&=\big\|\mathrm{offdiag}(\mathrm{Cov}(\bm{Z}))\big\|_F^2.
\end{align}
where $p_z(i)$ is the attention probability of latent $z$ on token $i$ and $\tau_{\mathrm{cov}}$ is an adaptive parameter, explained in detail in \suppsec{app:adaptive_cov}.
The total pretraining loss is
$\mathcal{L}_{\mathrm{MAE}}=\mathcal{L}_{\mathrm{rec}}
+\lambda_{\mathrm{cover}}\mathcal{L}_{\mathrm{cover}}
+\lambda_{\mathrm{feat}}\mathcal{L}_{\mathrm{feat}}$.

\myparagraph{Sparse Token Merger for Local Redundancy Reduction.} 
Before resampling, STM reduces local redundancy by merging tokens within non-overlapping $s\times s$ windows on the 2D grid.
For each window $g$ with index set $\mathcal{S}_g$, STM outputs $\tilde{\bm{x}}_g=\mathrm{LN}_{\mathrm{out}}\bigl(\frac{1}{\sqrt{|\mathcal{S}_g|}}
\sum_{i\in\mathcal{S}_g}\mathrm{LN}_{\mathrm{in}}(\bm{x}_i)\bigr)$,
$\tilde{\bm{c}}_g=(\lfloor c_i/s\rfloor,\lfloor r_i/s\rfloor)$, for any $i \in \mathcal{S}_g$, 
yielding merged tokens $\tilde{\bm{X}}\in\mathbb{R}^{N'\times D_v}$ with $N'\approx N/s^2$ and merged coordinates
$\tilde{\mathcal{C}}=\{\tilde{\bm{c}}_g\}$. We use $1/\sqrt{|\mathcal{S}_g|}$ (instead of $1/|\mathcal{S}_g|$)
for mixed-precision stability.
\vspace{-.3em}\subsection{Efficient Task-relevant Multimodal Fusion}
\label{sec:fusion}
Although the tile tokens are compressed into a compact set of $L$ latents, directly fusing all $L$ latents with the LLM in a standard LLaVA-style design is still costly and often unnecessary because many latents are task-irrelevant (Fig.~\ref{fig:cap_align_redundancy}).
We therefore adopt a \emph{select-then-fuse} principle: we first select only the Top-$M$ most query-relevant visual latents ($M \le L$), and then fuse these latents with the LLM via cross-attention. This reduces the fusion
  complexity from $O(TL)$ to $O(TM)$, where $T$ is the length of the text tokens.

\myparagraph{Token Importance Scorer (TIS).}
To further mimic human experts' query-dependent focus on different WSI regions and avoid spending compute on irrelevant evidence, TIS enables dynamic Top-$M$ latent selection. We project resampler latents to the LLM space as $\bm V=\mathrm{Proj}(\bm Z)\in\mathbb R^{L\times D}$ with rows $\bm v_i$. For sample $b$, we mean-pool text tokens to $\tilde{\bm q}_b\in\mathbb R^D$ and compute router scores
$\bm a_{b,i}=\mathrm{GELU}(W_v\bm v_i+W_q\tilde{\bm q}_b),\ \rho_{b,i}=\bm w^\top\bm a_{b,i}$.
We keep the ordered Top-$M$ list $\mathcal I_b=(i_{b,1},\ldots,i_{b,M})$ and route $\bm V_{\mathcal I_b}=[\bm v_{i_{b,1}},\ldots,\bm v_{i_{b,M}}]\in\mathbb R^{M\times D}$ to the LLM. TIS is trained by distilling CARA attention without extra labels. Let $\mathcal{S}_{\mathrm{CARA}}$ be the set of decoder layers where CARA is inserted, and let $H_{\text{head}}$ be the number of attention heads. At each layer $\ell\in\mathcal{S}_{\mathrm{CARA}}$, CARA performs text$\!\to\!$vision cross-attention over the $M$ routed latents $\bm{V}_{\mathcal{I}_b}$ and yields attention weights $\boldsymbol{\alpha}^{(\ell)} \in \mathbb{R}^{B \times H_{\text{head}} \times T \times M}$, where the last dimension indexes routed positions $m=1,\ldots,M$. We form a per-routed-latent teacher score by averaging over heads and text positions and then aggregating across CARA layers: $\bar{\alpha}_{b,m}=\tfrac{1}{|\mathcal{S}_{\mathrm{CARA}}|}\sum_{\ell\in\mathcal{S}_{\mathrm{CARA}}}\left(\tfrac{1}{H_{\text{head}}\,T}\sum_{h=1}^{H_{\text{head}}}\sum_{j=1}^{T}\alpha^{(\ell)}_{b,h,j,m}\right)$ for $b=1,\ldots,B$ and $m=1,\ldots,M$. 
We then form $\bm t_b=\mathrm{softmax}(\bar{\bm\alpha}_b/\tau_t)$ and $\bm p_b=\mathrm{softmax}(\bm\rho_{b,\mathcal I_b}/\tau_s)$, where $\bar{\bm\alpha}_b=[\bar{\alpha}_{b,1},\ldots,\bar{\alpha}_{b,M}]$ and $\bm\rho_{b,\mathcal I_b}=[\rho_{b,i_{b,1}},\ldots,\rho_{b,i_{b,M}}]$ and $\tau_t$ and $\tau_s$ are temperature terms. The TIS loss is the forward KL on routed latents only,
$\mathcal L_{\mathrm{TIS}}=\frac{1}{B}\sum_{b=1}^{B}\sum_{m=1}^{M} t_{b,m}\big(\log t_{b,m}-\log p_{b,m}\big)$,
plus a pairwise ranking term $\mathcal L_{\mathrm{rank}}$ that aligns \revblue{the relative TIS scores} with the teacher\revblue{, so that the Top-$M$ set selected by TIS agrees with the latents CARA attends to} (see \suppsec{app:rank_loss} for more details).\linelabel{rl:R11-method} For simplicity, we omit the sample subscript $b$ in the sequel.

\myparagraph{Cross-Attention Routing Adapter (CARA).}
CARA fuses the routed latents with the LLM through lightweight gated cross-attention for a small set of decoder layers $\mathcal{S}_{\mathrm{CARA}}$. Omitting the sample index $b$ for brevity, at each $\ell\in\mathcal S_{\mathrm{CARA}}$ we update hidden states as
$\tilde{\bm H}^{(\ell)}=\bm H^{(\ell)}+\gamma_\ell\,\mathrm{Attn}\!\big(\mathrm{LN}(\bm H^{(\ell)}),\bm V_{\mathcal I},\bm V_{\mathcal I}\big)$,
where $\gamma_\ell$ is a learnable gate initialized small for stable training. Moreover, cross-attention projection head is removed since the hidden dimension is aligned. Thus, the overall fusion objective is
$\mathcal L=\mathcal L_{\mathrm{LM}}+\lambda_{\mathrm{TIS}}\mathcal L_{\mathrm{TIS}}+\lambda_{\mathrm{rank}}\mathcal L_{\mathrm{rank}}$.

\vspace{-.5em}\section{Experiment}

\vspace{-.3em}\subsection{Experimental Settings}

\myparagraph{Dataset.}
We use three distinct datasets to evaluate our design choices. The WSI-Bench dataset, introduced in WSI-LLaVA~\cite{liang2025wsi}, is the primary source for training the resampler and for subsequent modality fusion with LLMs. It has over $9k$ WSIs and over $170k$ VQA pairs. For zero-shot testing, we assess the trained models on the WSI-Bench test set, as well as on Slide-Bench from SlideChat~\cite{chen2025slidechat} and selected data from Path-Bench~\cite{pathbench2024sun}. Notably, Slide-Bench contains BCNB WSIs~\cite{xu2021predicting}, which differ from the TCGA WSIs~\cite{weinstein2013cancer} in WSI-Bench, making it an effective out-of-distribution (OOD) test set. As both WSI-Bench and Path-Bench are based on the TCGA dataset, we remove any overlapping WSIs between WSI-Bench and Path-Bench training sets. After this adjustment, we retain 400 WSIs and 780 visual question-answer pairs from the Path-Bench dataset \revblue{for zero-shot evaluation}.
For the pre-processing, we first apply DSMIL~\cite{li2021dual} and CLAM~\cite{lu2021data} to identify tissue patches at $20\times$ magnification and remove background regions. By using the CONCH V1~\cite{lu2024avisionlanguage}, we then extract tile-level features from these patches. The tile size is $224\times224$. More details are provided in \suppsec{app:training_detail_train}.

\myparagraph{Experimental Details.}
The resampler is first pre-trained using MAE for $100k$ steps with $L=256$\revblue{; the $L{=}128$ variant uses a separately pretrained 128-latent resampler and its own Stage-2 checkpoint}.\linelabel{rl:R03-l128} After pre-training the resampler, we fuse the modality with the LLM in another two training stages. We use Qwen2.5-7B-Instruct~\cite{qwen25} as LLM. We insert CARA modules at decoder layers $\{1,3,5,7\}$. In the first fusion stage, we freeze the resampler and LLM, but with STM (\revblue{$s{=}2$}) on and only train the projector, TIS, and CARA with $M = 96$. During this stage, we train the model using the report data in WSI-Bench~\cite{liang2025wsi}. In the second fusion stage, we use all the VQA pairs and unfreeze the latents and the last layer of the resampler. Key hyperparameters are as follows. MAE pretraining uses 100k steps with \revblue{70}\% masking, AdamW (lr $10^{-4}$, 3\% warmup, cosine decay). Multimodal fusion stage-1 trains the projector+TIS+CARA (with $M{=}96$) for 3 epochs (global batch 512, lr $10^{-4}$) with the LLM and resampler frozen; stage-2 unfreezes the resampler last cross-attention layer and adds LoRA ($r{=}128$, $\alpha{=}64$) to LLM for 2 epochs (global batch \revblue{192}, lr $2\times10^{-5}$). We set $\lambda_{\mathrm{cover}}=5\times10^{-4}$, $\lambda_{\mathrm{feat}}=10^{-3}$, $\lambda_{\mathrm{TIS}}=0.02$\revblue{, and $\lambda_{\mathrm{rank}}=0.5$}. The loss weight sensitivity is studied in Sec.~\ref{sec:ablation}.  \revblue{MAE pretraining, Stage~1, and Stage~2 take 85.84, 4.49, and 117.20 GPU-hours on NVIDIA RTX 6000 Ada GPUs (207.53 in total), i.e., about 21.5, 1.1, and 29.3\,h on four GPUs.}\linelabel{rl:R04-cost} \revblue{All models are trained once with seed 42.}

\myparagraph{Baselines.} To verify the effectiveness of our LoC-Path framework, we compare it with several end-to-end MLLMs, including GPT-4o~\cite{hurst2024gpt}, Quilt-LLaVA~\cite{seyfioglu2024quilt}, WSI-LLaVA~\cite{liang2025wsi}, SlideChat~\cite{chen2025slidechat}\revblue{, MLLM-HWSI~\cite{alawode2026mllmhwsi}, and TCP-LLaVA~\cite{lyu2025efficient}, plus a no-compression reference feeding up to $20{,}480$ raw tile tokens into the same Qwen2.5-7B}. The input tokens for SlideChat~\cite{chen2025slidechat} are $20480$. We resize the WSI thumbnails to $1024\times1024$ for Quilt-LLaVA~\cite{seyfioglu2024quilt}. We evaluate representative token reduction methods, including ToMe~\cite{bolya2023token,zhong2025aim}, DivPrune~\cite{alvar2025divprune}, ACMIL~\cite{zhang2024attention}, and adaptive average pooling, as drop-in replacements for our STM+resampler. For a fair comparison, we keep the projector/TIS/CARA modules and the two-stage training recipe unchanged, and match the token budget by reducing each slide to $256$ visual tokens \revblue{($L{=}256$)}, from which TIS routes $M=96$ tokens to the LLM \revblue{(Tab.~\ref{tab:main_results_wsi_bench})}.\linelabel{rl:R03-budget} We only tune the reduction ratio to meet the target budget; full details are provided in \suppsec{app:training_detail_eval}. 

\myparagraph{BRCA MIL Tasks.}
We compared resampler with slide-level encoders \cite{ding2025titan,shaikovski2024prism,xu2024gigapath} on TCGA-BRCA with \textbf{patient-level} splits on (i) Stage classification (Stage~I vs.\ Others; unknown removed) and (ii) Vital Status (alive vs.\ deceased). Compared slide-level encoders use their default pretrained weights and settings. All methods use the same ABMIL head~\cite{ilse2018attention} with lr $10^{-4}$; encoders are frozen, and only the head is trained, selecting checkpoints by the best validation AUC. Full dataset statistics and splits are in \suppsec{app:training_detail_eval}.

\myparagraph{Evaluation Metrics.}
For the \revblue{VQA} tasks, we follow the approach of WSI-LLaVA~\cite{liang2025wsi} and categorize the tasks into four types: morphological analysis, diagnosis, treatment planning, and report generation. For open-ended questions, we evaluate performance using BLEU~\cite{papineni2002bleu}, WSI-Precision (WSI-P), and WSI-Relevance (WSI-R), while accuracy is used for closed-ended tasks. WSI-P and WSI-R are GPT-4o-based~\cite{hurst2024gpt} evaluation metrics, proposed in WSI-LLaVA~\cite{liang2025wsi}. We follow WSI-LLaVA~\cite{liang2025wsi}'s exact prompts and scoring rubric; details are in \suppsec{app:wsi_metrics}. To assess the effectiveness of our resampler and the quality of the slide-level latent representations, we use AUC~\cite{hanley1982meaning} and Accuracy~\cite{stork2001pattern} as the primary metrics for classification tasks. For the feature quality assessment from the resampler, we assess the latent attention coverage at a fixed threshold $10^{-4}$ with $\mathbb{E}\big[\frac{1}{N_v}\sum_i \mathbf{1}\{\max_z p_z(i)>10^{-4}\}\big]$, where $\mathbb{E}[\cdot]$ denotes the mean over evaluation WSIs and $N_v$ is the number of valid tile tokens per WSI (with $u(i)=\max_z p_z(i)$ in Eq.~\ref{eq:cover}); we also report the \revblue{mean} pairwise cosine similarity between latents \revblue{(lower means more dispersed latents)}. \revblue{The supplementary provides redundancy analyses (Sec.~\ref{app:redundancy}), method details (ranking loss, CARA complexity; Sec.~\ref{app:method_details}), implementation and training cost (Sec.~\ref{app:training_detail}), baseline and WSI-P/WSI-R protocols (Sec.~\ref{app:training_detail_eval}), extended ablations and pretraining comparisons (Sec.~\ref{app:more_ablation}), and Fig.~\ref{fig:tis_attention} construction with qualitative results and failure cases (Sec.~\ref{app:more_qualitative_results}).}\linelabel{rl:H22-supp} 

\begin{table*}[t]
  \caption{\textbf{Detailed performance and efficiency comparison} by task type, together with report generation and zero-shot generalization (Slide-Bench and Path-Bench subset).
  WSI-P/WSI-R denote WSI-Precision/Relevance and Acc denotes Accuracy;
  $L$ \revblue{is the visual-token length after reduction or resampling (raw length if none)} and $M$ \revblue{is} the number routed to the LLM.
  Mem.\ (GB) and TFLOPs are per forward pass for LLM generation only (visual inputs pre-extracted for SlideChat and WSI-LLaVA; token-reduction baselines share LoC-Path's $L{=}256$ LLM-side path)\revblue{; SlideChat is re-measured at its full $N{=}20{,}480$-token input}.
  Clin.\ Avg is the mean of WSI-P and WSI-R for report generation.
  The last column is $\frac{1}{2}\big(\mathrm{VQA\ Avg}+\mathrm{Clin.\ Avg}\big)$.
  Best results per column are in bold.
  $^{*}$\,denotes WSI-LLaVA reproduced under our unified evaluation pipeline (token-budget variants in \supptab{tab:wsi_llava_token_pooling}). \revblue{Details of MLLM-HWSI, TCP-LLaVA, and the raw-token row are in \suppsec{app:r2_baselines}.}}
  \vspace{-.6em}
\centering
\scriptsize
\setlength{\tabcolsep}{2pt}
\renewcommand{\arraystretch}{0.75}
\resizebox{\textwidth}{!}{%
\begin{tabular}{@{}lccrr ccc ccc ccc c ccc cccc cc c@{}}
\toprule
\multirow{3}{*}{\textbf{Model}} &
\multicolumn{4}{c}{\textbf{Cost}} &
\multicolumn{10}{c}{\textbf{WSI-Bench: VQA}} &
\multicolumn{7}{c}{\textbf{WSI-Bench: Report Gen.}} &
\multicolumn{2}{c}{\textbf{Zero-shot}} &
\multirow{3}{*}{\makecell[c]{\textbf{VQA+}\\\textbf{clin.}$\uparrow$}} \\
\cmidrule(lr){2-5}\cmidrule(lr){6-15}\cmidrule(lr){16-22}\cmidrule(lr){23-24}
 & \multirow{2}{*}{$L$} & \multirow{2}{*}{$M$} &
\multirow{2}{*}{\makecell[c]{Mem.\\(GB)$\downarrow$}} &
\multirow{2}{*}{TFLOPs$\downarrow$} &
\multicolumn{3}{c}{Morph.} &
\multicolumn{3}{c}{Diagnosis} &
\multicolumn{3}{c}{Treatment} &
\multirow{2}{*}{Avg.} &
\multicolumn{3}{c}{Clin.} &
\multicolumn{4}{c}{NLU} &
Slide-Bench &
Path-Bench & \\
\cmidrule(lr){6-8}\cmidrule(lr){9-11}\cmidrule(lr){12-14}\cmidrule(lr){16-18}\cmidrule(lr){19-22}\cmidrule(lr){23-23}\cmidrule(lr){24-24}
 & & & & & WSI-P & WSI-R & Acc & WSI-P & WSI-R & Acc & WSI-P & WSI-R & Acc & & WSI-P & WSI-R & Avg & BLEU-1 & BLEU-2 & ROUGE-L & METEOR & Acc & Acc & \\
\midrule
\multicolumn{25}{l}{\emph{General MLLMs}} \\
GPT-4o~\cite{hurst2024gpt} & - & - & - & - & 0.220 & 0.204 & 0.471 & 0.472 & 0.457 & 0.530 & 0.496 & 0.841 & 0.875 & 0.507 & 0.067 & 0.138 & 0.103 & 0.202 & 0.069 & 0.132 & 0.167 & 0.414 & 0.793 & 0.305 \\
Quilt-LLaVA~\cite{seyfioglu2024quilt} & 576 & 576 & 12.620 & 12.798 & 0.448 & 0.447 & 0.947 & 0.586 & 0.604 & 0.849 & 0.788 & 0.816 & \revblue{\textbf{1.000}} & 0.721 & 0.324 & 0.333 & 0.329 & 0.474 & 0.351 & 0.475 & 0.460 & 0.163 & 0.380 & 0.525 \\
SlideChat~\cite{chen2025slidechat} & 20480 & 20480 & 11.583 & \revblue{267.672} & 0.269 & 0.281 & 0.870 & 0.319 & 0.385 & 0.767 & 0.654 & 0.577 & 0.333 & 0.495 & 0.123 & 0.168 & 0.146 & 0.439 & 0.107 & 0.161 & 0.137 & 0.546 & \textbf{0.847} & 0.320 \\
\revblue{MLLM-HWSI~\cite{alawode2026mllmhwsi}} & \revblue{64} & \revblue{64} & \revblue{-} & \revblue{-} & \revblue{0.457} & \revblue{0.517} & \revblue{0.886} & \revblue{0.370} & \revblue{0.409} & \revblue{0.731} & \revblue{0.613} & \revblue{0.664} & \revblue{0.812} & \revblue{0.606} & \revblue{0.153} & \revblue{0.244} & \revblue{0.198} & \revblue{0.308} & \revblue{0.215} & \revblue{0.315} & \revblue{0.320} & \revblue{-} & \revblue{-} & \revblue{0.402} \\
\revblue{TCP-LLaVA~\cite{lyu2025efficient}} & \revblue{2000} & \revblue{2000} & \revblue{-} & \revblue{-} & \revblue{0.270} & \revblue{0.268} & \revblue{0.875} & \revblue{0.350} & \revblue{0.386} & \revblue{0.749} & \revblue{0.661} & \revblue{0.580} & \revblue{0.354} & \revblue{0.499} & \revblue{0.109} & \revblue{0.152} & \revblue{0.131} & \revblue{0.097} & \revblue{0.037} & \revblue{0.159} & \revblue{0.109} & \revblue{-} & \revblue{-} & \revblue{0.315} \\
WSI-LLaVA~\cite{liang2025wsi} & \revblue{576} & 576 & 13.619 & 10.491 & 0.488 & 0.610 & \revblue{\textbf{0.951}} & \revblue{\textbf{0.610}} & 0.612 & \revblue{\textbf{0.863}} & \revblue{\textbf{0.810}} & \revblue{\textbf{0.845}} & \revblue{\textbf{1.000}} & \textbf{0.754} & 0.380 & 0.429 & 0.405 & 0.480 & \revblue{\textbf{0.358}} & \revblue{\textbf{0.490}} & \revblue{\textbf{0.465}} & \textbf{0.553} & - & 0.579 \\
WSI-LLaVA$^{*}$~\cite{liang2025wsi} & \revblue{576} & 576 & 13.619 & 10.491 & \revblue{\textbf{0.567}} & 0.605 & 0.926 & 0.594 & \revblue{\textbf{0.639}} & 0.826 & 0.774 & 0.786 & \revblue{\textbf{1.000}} & 0.746 & 0.383 & 0.437 & 0.410 & 0.418 & 0.306 & 0.410 & 0.415 & - & - & 0.578 \\
\midrule
\multicolumn{25}{l}{\emph{\revblue{No compression}}} \\
\revblue{Qwen2.5-7B (raw tokens)} & \revblue{20480} & \revblue{20480} & \revblue{-} & \revblue{-} & \revblue{0.561} & \revblue{0.601} & \revblue{0.941} & \revblue{0.556} & \revblue{0.600} & \revblue{0.854} & \revblue{0.731} & \revblue{0.747} & \revblue{\textbf{1.000}} & \revblue{0.732} & \revblue{0.372} & \revblue{0.430} & \revblue{0.401} & \revblue{0.409} & \revblue{0.299} & \revblue{0.408} & \revblue{0.406} & \revblue{-} & \revblue{-} & \revblue{0.567} \\
\midrule
\multicolumn{25}{l}{\emph{Token reduction baselines}} \\
ToMe (AIM)~\cite{zhong2025aim} & 256 & \revblue{96} & \textbf{7.709} & 3.388 & 0.535 & 0.593 & 0.942 & 0.528 & 0.556 & 0.842 & 0.733 & 0.779 & \revblue{\textbf{1.000}} & 0.723 & 0.377 & 0.469 & 0.423 & \textbf{0.586} & 0.309 & 0.384 & 0.418 & 0.508 & 0.820 & 0.573 \\
DivPrune~\cite{alvar2025divprune} & 256 & \revblue{96} & \textbf{7.709} & 3.388 & 0.527 & 0.581 & 0.921 & 0.491 & 0.515 & 0.820 & 0.518 & 0.627 & 0.854 & 0.650 & 0.335 & 0.419 & 0.377 & 0.556 & 0.286 & 0.375 & 0.417 & 0.544 & 0.737 & 0.514 \\
ACMIL~\cite{zhang2024attention} & 256 & \revblue{96} & \textbf{7.709} & 3.388 & 0.535 & 0.593 & 0.943 & 0.528 & 0.556 & 0.842 & 0.733 & 0.779 & \revblue{\textbf{1.000}} & 0.723 & 0.334 & 0.422 & 0.378 & 0.572 & 0.294 & 0.374 & 0.401 & 0.508 & 0.821 & 0.550 \\
Avg.\ Pooling & 256 & \revblue{96} & \textbf{7.709} & 3.388 & 0.532 & 0.593 & 0.940 & 0.545 & 0.574 & 0.845 & 0.666 & 0.766 & 0.958 & 0.713 & 0.320 & 0.423 & 0.372 & 0.569 & 0.296 & 0.372 & 0.406 & 0.545 & 0.837 & 0.542 \\
\midrule
\multicolumn{25}{l}{\emph{No resampler pretraining baselines}} \\
LLM-loss~\cite{liu2023llava} & 256 & 96 & \revblue{\textbf{7.709}} & 3.388 & 0.467 & 0.540 & 0.921 & 0.425 & 0.452 & 0.784 & 0.710 & 0.759 & 0.958 & 0.668 & 0.148 & 0.249 & 0.199 & 0.327 & 0.229 & 0.329 & 0.329 & 0.447 & 0.806 & 0.433 \\
\midrule
\textcolor{OursRed}{LoC-Path} & 256 & 96 & \textbf{7.709} & 3.388 & 0.564 & \textbf{0.626} & 0.943 & 0.570 & 0.598 & 0.860 & 0.734 & 0.771 & 0.979 & 0.738 & 0.369 & 0.462 & 0.416 & 0.579 & 0.312 & 0.390 & \revblue{0.426} & 0.519 & 0.813 & 0.577 \\
\textcolor{OursRed}{LoC-Path} & 128 & 96 & \textbf{7.709} & \textbf{3.019} & \textbf{0.567} & 0.620 & 0.931 & \revblue{0.582} & \revblue{0.613} & 0.849 & 0.733 & 0.769 & \textbf{1.000} & \revblue{0.740} & \textbf{0.384} & \textbf{0.473} & \textbf{0.429} & 0.583 & \revblue{0.313} & \revblue{0.392} & 0.425 & 0.519 & \textbf{0.847} & \textbf{0.584} \\
\bottomrule
\end{tabular}
}
\label{tab:main_results_wsi_bench}
\vspace{-1.5em}
\end{table*}

\begin{figure}
    \centering
\includegraphics[width=1.0\linewidth]{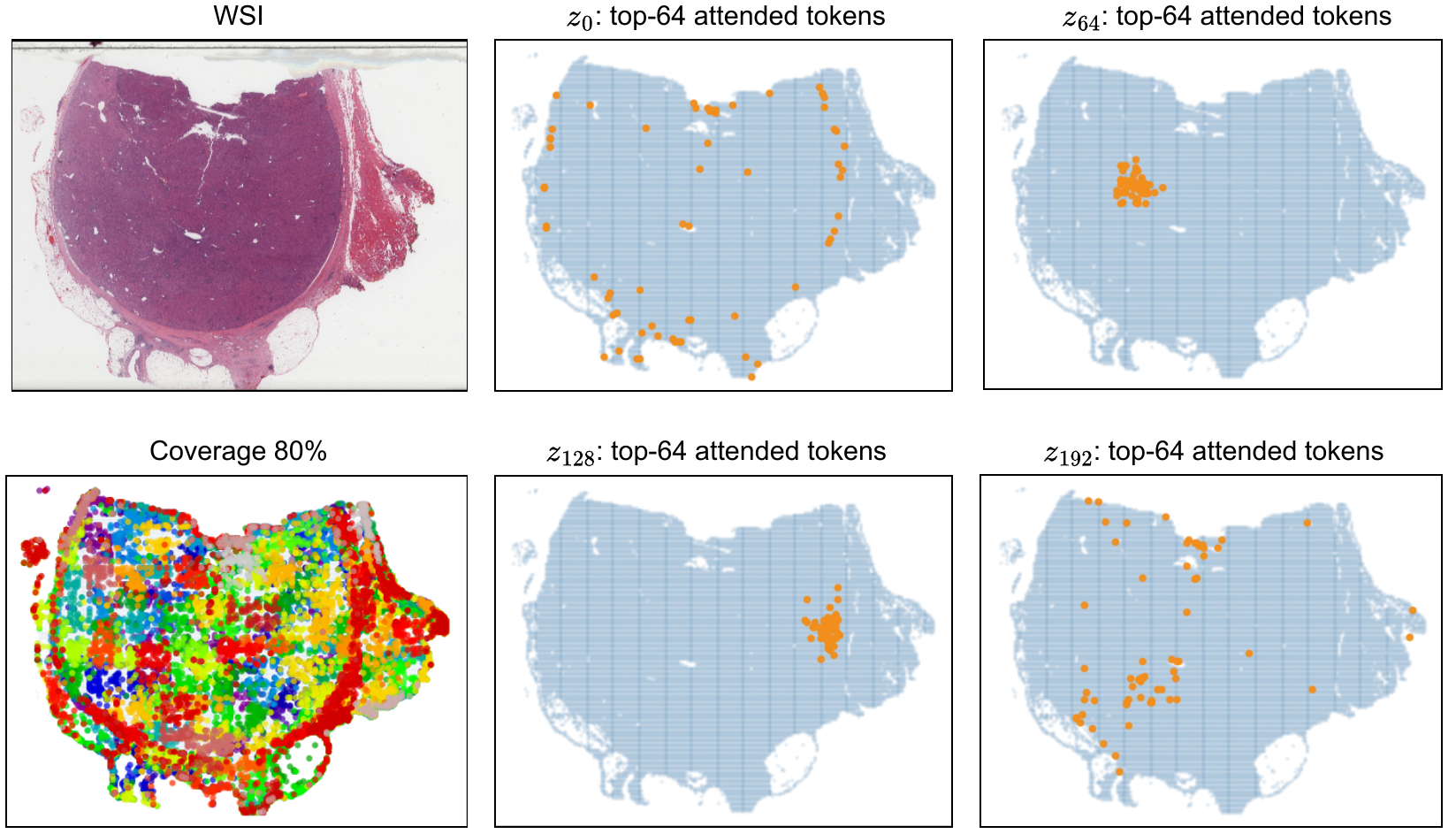}
\vspace{-2em}
    \caption{Coverage and visualization of latent attention preferences. Bottom-left: attention coverage at threshold $10^{-4}$; 80\% coverage means the resampler attends to 80\% of token positions on the slide. $z_0$ shows the top-64 token locations attended by latent $z_0$ (orange); blue indicates all token positions.
    }
\vspace{-1.6em}
    \label{fig:covery_and_attention_map}
\end{figure}

\vspace{-.3em}\subsection{Results}
\label{sec:results}
\myparagraph{Slide-level MLLM Performance.}
In Tab.~\ref{tab:main_results_wsi_bench}, LoC-Path and WSI-LLaVA are closely matched on VQA, especially under our unified evaluation pipeline. \revblue{A paired bootstrap on the 3,194 questions evaluable for both models (\suppsec{app:r2_stats}) gives a LoC-Path minus WSI-LLaVA difference of $-0.010$ on the VQA average (95\% CI $[-0.023, 0.003]$) and $-0.002$ on VQA+clinical (95\% CI $[-0.020, 0.016]$); both intervals include zero, so the two models are not statistically distinguishable on this subset.}\linelabel{rl:R06-ci} On the other hand, LoC-Path \revblue{scores higher} on clinical report generation 
(Clin.\ Avg 0.429 \textit{vs.}\ 0.410).
LoC-Path also matches the best zero-shot Path-Bench accuracy (0.847) 
with a modest trade-off on Slide-Bench.
Compared with the token reduction baselines that replace our 
STM+resampler with alternative compression strategies under the 
same fusion pipeline, LoC-Path \revblue{obtains the higher VQA and combined averages} 
(e.g., VQA Avg \revblue{0.740} \textit{vs.}\ 0.723 for ToMe/ACMIL), 
\revblue{supporting} the advantage of learning to compress over 
heuristic reduction. \revblue{Feeding up to $20{,}480$ raw tile tokens per slide (uniformly subsampled beyond that) to the same Qwen2.5-7B without compression (Tab.~\ref{tab:main_results_wsi_bench}) gives 0.732/0.401 VQA/Clin.\ Avg, so compressing to $M{=}96$ routed latents does not hurt accuracy in this setting, while its training is estimated at 357 GPU-hours, 1.7$\times$ the 207.53 of all three LoC-Path stages (\supptab{tab:r2_training_cost}).}\linelabel{rl:R17-raw} \revblue{Comparable performance to} WSI-LLaVA is achieved with much fewer visual tokens sent to the LLM 
(96 \textit{vs.}\ 576)\revblue{; all Tab.~\ref{tab:main_results_wsi_bench} entries are single runs, read as point estimates rather than statistical superiority}.

\begin{table*}[t]
\caption{Patient-level MIL results on TCGA-BRCA (Stage, Vital Status), with feature quality and computational cost. Feature quality is measured on $2k$ WSIs sampled from TCGA based on tumor type. \revblue{Feature diversity is the mean pairwise cosine among output latents (lower is more diverse; std across WSIs).} Computational cost (peak reserved memory and TFLOPs) is measured during inference under max tile token number for each encoder; "---" denotes not applicable. Resampler architecture ablations are in \supptab{tab:ablation_resampler}.}
\vspace{-.8em}
\centering
\scriptsize
\setlength{\tabcolsep}{8pt}
\renewcommand{\arraystretch}{0.75}
\begin{tabular}{@{}l cc cc cc cc@{}}
\toprule
\multirow{2}{*}{\textbf{Model}} &
\multicolumn{2}{c}{\textbf{Stage (I vs Others)}} &
\multicolumn{2}{c}{\textbf{Vital Status}} &
\multicolumn{2}{c}{\textbf{Feature Quality (mean)}} &
\multicolumn{2}{c}{\textbf{Computational Cost}} \\
\cmidrule(lr){2-3} \cmidrule(lr){4-5} \cmidrule(lr){6-7} \cmidrule(lr){8-9}
 & {AUC $\uparrow$} & {Acc $\uparrow$} & {AUC $\uparrow$} & {Acc $\uparrow$} & {\revblue{Feature diversity} (std) $\downarrow$} & {Coverage $\uparrow$} & {Peak Res.\ (GB) $\downarrow$} & {TFLOPs $\downarrow$} \\
\midrule
\multicolumn{9}{l}{\textit{Slide-level encoders}} \\
TITAN~\cite{ding2025titan} & \textbf{0.836} & \textbf{0.781} & 0.718 & 0.863 & 0.372 (0.189) & --- & 9.27 & \textbf{0.350} \\
PRISM~\cite{shaikovski2024prism} & 0.737 & 0.730 & 0.643 & 0.869 & 0.782 (0.226) & --- & 7.90 & 3.530 \\
GigaPath~\cite{xu2024gigapath} & 0.728 & 0.688 & 0.709 & 0.845 & 0.382 (0.197) & --- & 4.02 & 10.340 \\
COBRA~\cite{lenz2025cobra} & 0.772 & 0.750 & 0.642 & 0.851 & 0.722 (0.127) & --- & 2.73 & 2.611 \\
\midrule
\multicolumn{9}{l}{\textit{Resampler pre-training / regularization (ours)}} \\
LLM-loss~\cite{liu2023llava} & 0.724 & 0.688 & \textbf{0.759} & \textbf{0.882} & 0.841 (0.049) & \textbf{1.000} & \textbf{2.25} & 1.713\\
Ours (no regularizer) & 0.741 & 0.672 & 0.714 & 0.845 & \textbf{0.110} (0.462) & 0.812 & \textbf{2.25} & 1.713 \\
\textbf{Ours (MAE)} & 0.763 & 0.734 & 0.696 & 0.845 & 0.346 (0.056) & 0.894 & \textbf{2.25} & 1.713 \\
\bottomrule
\end{tabular}
\label{tab:resampler_comparison}
\vspace{-2.0em}
\end{table*}

\myparagraph{Efficiency Where It Matters.}
Tab.~\ref{tab:fusion} and Fig.~\ref{fig:latency_scaling} summarize the fusion ablation and deployment-time scaling: LoC-Path keeps time-to-first-token (TTFT) and additional prefill memory nearly constant as visual context grows, while SlideChat~\cite{chen2025slidechat} scales sharply.
Compared with other LLaVA-style slide-level MLLMs using 7B LLMs (Qwen2.5-7B~\cite{qwen25} or WSI-LLaVA's Vicuna-7B~\cite{zheng2023judging}), our model is the most computationally efficient in Tab.~\ref{tab:main_results_wsi_bench}. The inference TFLOPs of our model ($L=128/M=96$) are $\revblue{96.9}\%$ lower than the mean of all General baselines, and inference memory is $38.9\%$ lower. \revblue{Tab.~\ref{tab:main_results_wsi_bench} counts LLM generation only; adding the resampler's cost at the maximum tile budget (1.713 TFLOPs, Tab.~\ref{tab:resampler_comparison}) gives 5.101 TFLOPs for the full LoC-Path path ($L{=}256$), still far below the 267.672 TFLOPs that SlideChat spends on LLM generation alone.} The token-reduction baselines spend at most 0.025 TFLOPs on their visual stage and share the same $M{=}96$ LLM path, so their cost matches LoC-Path and they differ mainly in accuracy.\linelabel{rl:R05-cost}

\begin{figure}[t]
\centering
\includegraphics[width=0.444\linewidth]{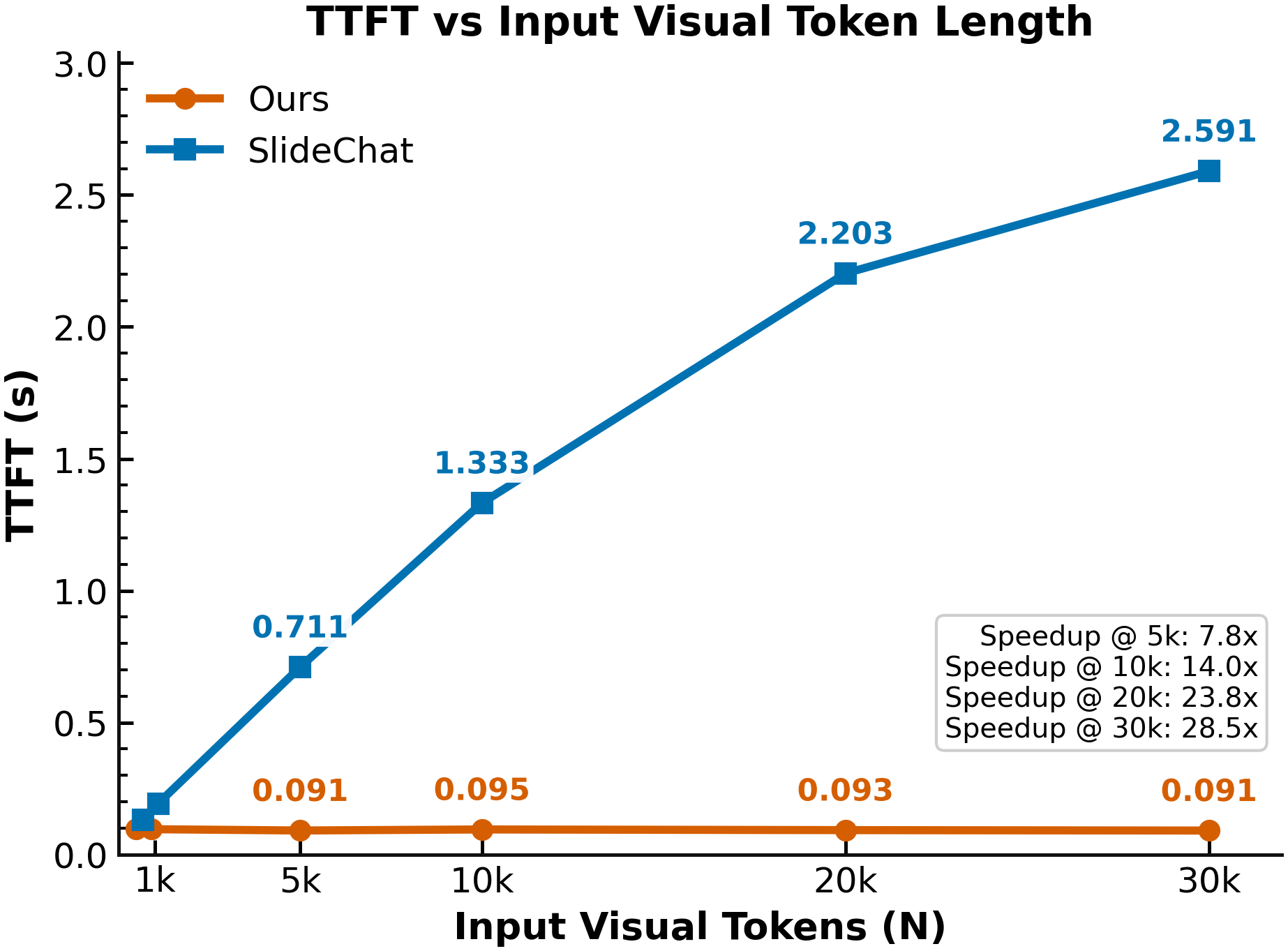}\hfill
\includegraphics[width=0.476\linewidth]{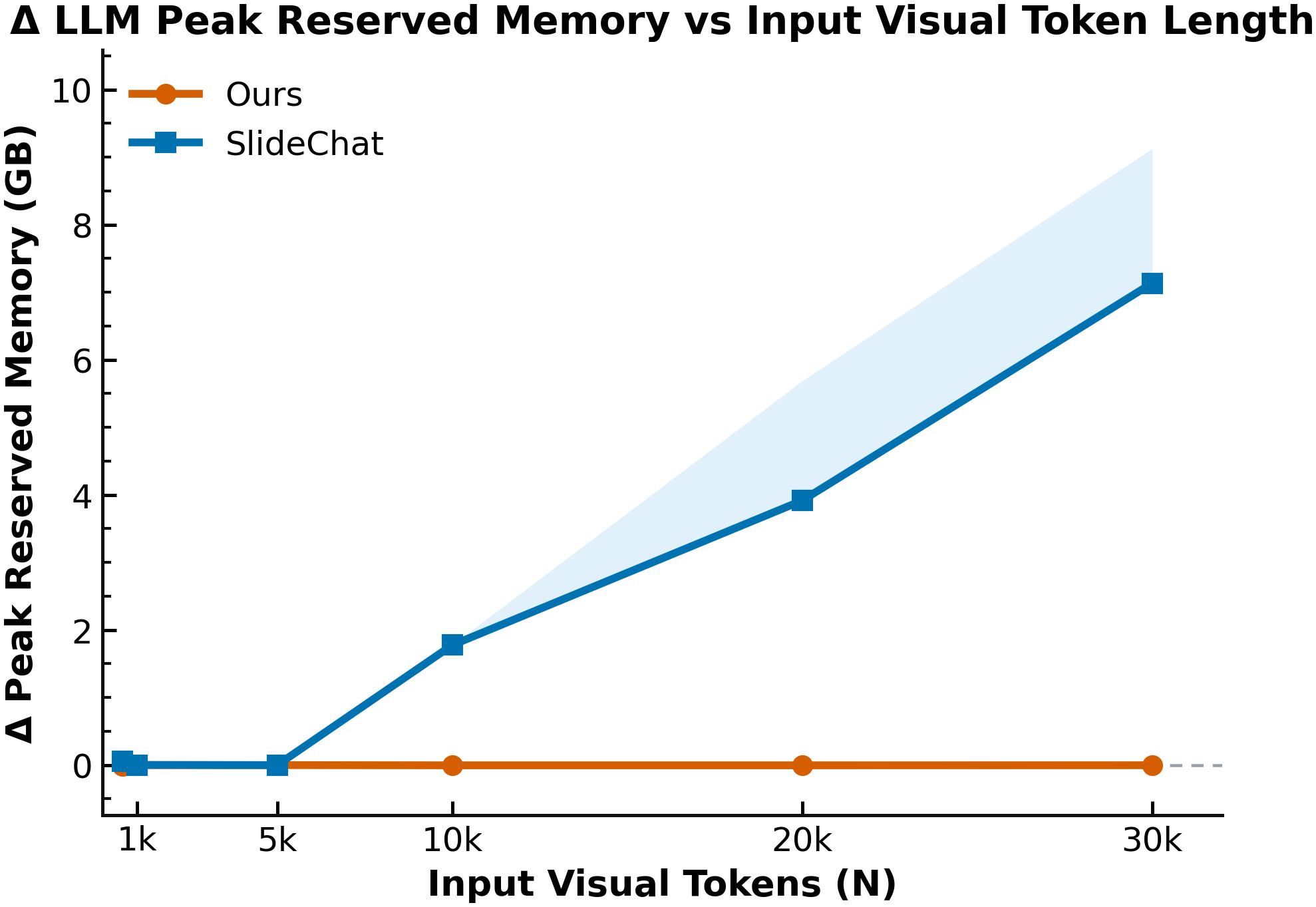}
\vspace{-1.0em}
\caption{\textbf{Deployment scaling.} Inference latency (TTFT) and $\Delta$ peak reserved memory \textit{vs.}\ input visual token length $N$; LoC-Path stays nearly constant while SlideChat grows with visual context.}
\vspace{-1.7em}
\label{fig:latency_scaling}
\end{figure}

\myparagraph{Resampler as Lightweight Slide-level Encoder.} To test whether aggressive compression preserves clinically relevant signals, we compare against representative slide-level encoders (TITAN~\cite{ding2025titan}, PRISM~\cite{shaikovski2024prism}, COBRA~\cite{lenz2025cobra} and GigaPath~\cite{xu2024gigapath}) on the two MIL tasks above. As shown in Tab.~\ref{tab:resampler_comparison}, the resampler achieves competitive classification performance, though not uniformly (\eg, TITAN and COBRA lead on Stage), while using the lowest memory and fewer TFLOPs than most slide-level encoders.\linelabel{rl:R13-cindex} The compared slide-level encoders are pretrained on much larger private datasets than the 9$k$ WSI-Bench~\cite{liang2025wsi} slides used for our resampler, which likely explains the remaining gap. Fig.~\ref{fig:covery_and_attention_map} visualizes latent coverage and diversity.

\begin{figure}
    \centering
    \includegraphics[width=1.0\linewidth]{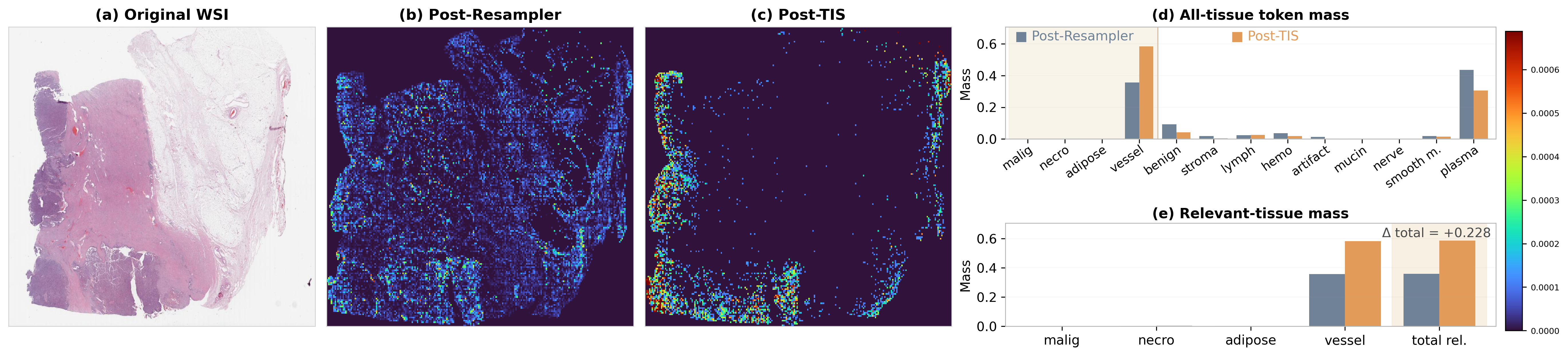}
    \vspace{-2em}
    \caption{Post-hoc interpretation of TIS. (a) WSI; (b) Post-Resampler and (c) Post-TIS patch-mass maps (warmer = higher mass); (d) mass over 13 weak tissue labels; (e) mass over the report-relevant classes. Here TIS raises the relevant mass by $+0.228$, mainly via blood-vessel enrichment. Labels are used only for interpretation (\suppsec{app:tissue_labeling}).}
    \label{fig:tis_attention}
    \vspace{-1.7em}
\end{figure}

\vspace{-.3em}\subsection{Ablation Study}
\vspace{-.3em}
\label{sec:ablation}

\myparagraph{Token Compression Preserving Small Region.}
\label{app:lesion_recall}
We stress-test this on a CAMELYON16~\cite{bejnordi2017camelyon} focal-lesion subset---$31$ slides whose expert-annotated tumor covers $<2\%$ of tissue \emph{without \revblue{fine-tuning any LoC-Path module}}. \revblue{We trace information preservation from STM through the resampler. STM reduces the token count by 73.8\% while the FROC of a linear lesion probe improves from 0.624 to 0.744; after resampling and $M{=}96$ routing, 98.75\% of the lesion tiles remain covered. Lesion location is not linearly recoverable from the latents (probe FROC 0.065), so compression is lossy; what is preserved is that lesion tiles stay within the resampler's attention support (\suppsec{app:r2_lesion}).}\linelabel{rl:R08-lesion}

\myparagraph{Design Rationale of the MAE-pretrained Resampler.}
We show that MAE pre-training is necessary: compared to end-to-end training with an LLM, MAE pre-training prevents latent features from collapsing to a global average (Tab.~\ref{tab:resampler_comparison}). We further prove this in Tab.~\ref{tab:main_results_wsi_bench}, where simple training resampler with LLM-loss only hurts downstream performance by a large margin. The two regularization terms \revblue{keep latent diversity stable across slides (the std of the mean pairwise cosine drops from $0.462$ to $0.056$)} and increase attention coverage over tile tokens \revblue{($0.812\rightarrow0.894$) in} Tab.~\ref{tab:resampler_comparison}.\linelabel{rl:R09-div}  Resampler with dense attention tends to learn average features, which \revblue{lowers} the model's performance ($0.738\rightarrow0.655$) in \supptab{tab:detailed_ablation}. This explains the necessity of Top-$K$ attention. 

\begin{wraptable}{r}{0.45\columnwidth}
\centering
\vspace{-1.6em}
\caption{\textbf{Fusion ablation.} WSI-Bench Avg (excluding report generation) across fusion designs.}
\vspace{-.8em}
\label{tab:fusion}
\scriptsize
\setlength{\tabcolsep}{5pt}
\renewcommand{\arraystretch}{0.85}
\begin{tabular}{@{}lc@{}}
\toprule
Fusion & Avg $\uparrow$ \\
\midrule
LLaVA (concat) & 0.733 \\
Flamingo (cross-attn) & 0.726 \\
Flamingo + TIS & 0.729 \\
Ours w/o TIS \revblue{(no routing)} & 0.736 \\
Ours w/o $\mathcal{L}_{\mathrm{rank}}$ & 0.728 \\
Ours & \textbf{0.738} \\
\bottomrule
\end{tabular}
\end{wraptable}
\myparagraph{Design Rationale of TIS and CARA Modules.} Tab.~\ref{tab:fusion} decomposes the benefit of our routed fusion design. Adding TIS to a standard Flamingo-style cross-attention baseline improves the WSI-Bench average from $0.726$ to $0.729$\revblue{, a small gain}. Within our full CARA+TIS model, removing TIS \revblue{(all $256$ latents exposed, no routing)} reduces the average from $0.738$ to $0.736$\revblue{. At the same $M{=}96$ budget, TIS improves BCNB out-of-distribution accuracy over simple average pooling of the no-TIS model (0.521 vs.\ 0.496; \suppsec{app:r2_tis})}.\linelabel{rl:R10-tis} Removing the TIS ranking loss causes a larger drop, from $0.738$ to $0.728$\revblue{. Note that CARA does not depend on the order of the routed latents, so this gain is not due to ordering; rather, the pairwise supervision sharpens the TIS scores so that the Top-$M$ set it routes better matches the latents that CARA relies on}.\linelabel{rl:R11-rank} Fig.~\ref{fig:tis_attention} shows a post-hoc case in which TIS reallocates patch mass toward report-derived tissue categories. Additional ablations in \supptab{tab:detailed_ablation} further show that multi-layer CARA insertion is more effective than a single early-layer CARA module\revblue{, and that the learnable CARA gate outperforms fixing $\gamma_\ell=1$}. 

\myparagraph{Loss Weight Sensitivity.}
We assess \revblue{the effect of} the two fusion-loss weights $\lambda_{\mathrm{TIS}}$ and $\lambda_{\mathrm{rank}}$ \revblue{(on the $L{=}256$ model)} on \revblue{WSI-Bench and} the BCNB out-of-distribution test. \revblue{For $\lambda_{\mathrm{TIS}}\in\{0.01,0.02,0.05\}$, the WSI-Bench closed-task macro accuracy ranges from $0.854$ to $0.927$ and BCNB accuracy from $0.476$ to $0.519$, with $0.02$ giving the best result; for $\lambda_{\mathrm{rank}}\in\{0,0.05,0.5,5\}$, the corresponding ranges are $0.915$--$0.927$ and $0.502$--$0.519$. Thus $\lambda_{\mathrm{rank}}$ is stable over the tested range, whereas $\lambda_{\mathrm{TIS}}$ has a clearer effect (full results in \suppsec{app:r2_sens}); BCNB and Path-Bench are used only for reporting, not for model or hyperparameter selection.}\linelabel{rl:R14-sens} 

\begin{wrapfigure}{r}{0.42\columnwidth}
\centering
\vspace{-1.4em}
\includegraphics[width=0.40\columnwidth]{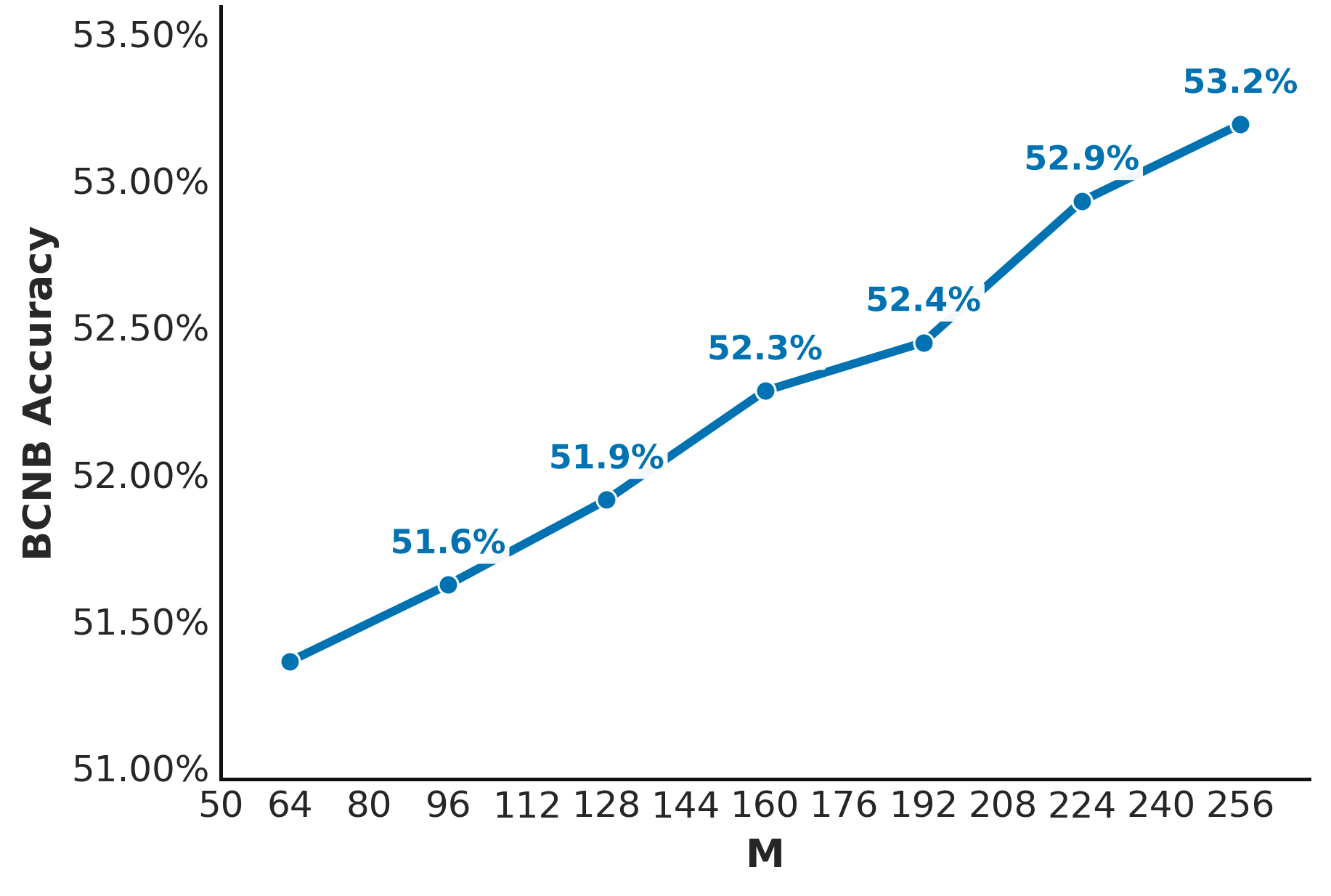}
\vspace{-.5em}
\caption{Routing budget on Slide-Bench BCNB.}
\label{fig:bcnb_acc_vs_m}
\vspace{-1.4em}
\end{wrapfigure}
\myparagraph{Impact of Routed Token Budget.} Increasing the selected token budget $M$ steadily while fixing latent length $L$ improves OOD accuracy as shown in Fig.~\ref{fig:bcnb_acc_vs_m}. The full $(L,M)$ sweeps on WSI-Bench are reported in \supptab{tab:ablation_compact_singlecol}.

\vspace{-0.5em}
\section{Conclusion}
\vspace{-0.3em}

We propose LoC-Path, an efficient slide-level MLLM motivated by the strong redundancy of WSI tile tokens and their sparse, query-dependent task relevance under slide-level language supervision. It compresses before fusion: STM merges locally redundant tiles, an MAE-pretrained resampler compresses the slide into a compact latent set, and TIS+CARA route only the Top-$M$ query-relevant latents into a few LLM layers. LoC-Path enables slide-level MLLM development and deployment under limited computational resources while remaining competitive with prior slide-level MLLMs.

\myparagraph{\revblue{Limitations and Future Work.}} \revblue{LoC-Path depends on a fixed tile encoder, so domain shift in the encoder propagates to the LLM. Its compression favors global coverage over worst-case preservation, so subtle lesions can be under-represented, and generated reports can be globally plausible while missing fine-grained details (e.g., grades or biomarker status) that LLM-judge metrics only partly penalize. Future work will couple the encoder, compression, and routing modules under objectives that preserve rare but critical patterns, adapt the token budgets to the question and slide, and validate on multi-institutional data with pathologist evaluation and calibrated uncertainty.}\linelabel{rl:R12-limit}
\label{sec:main_end}

\subsubsection*{Acknowledgments}
This work was partially supported by grants NSF CCF-2144901, NIH R01NS143143, and R01CA297843. Jiawei Zhou is supported by an Amazon Research Award and a Coefficient Giving Grant.

{
    \small
    \bibliographystyle{ieeenat_fullname}
    \bibliography{main}
}

\fi  %

\ifbuildsupp
  \ifnum\BuildMode=1\relax
    \clearpage
  \fi
  \ifnum\BuildMode=2\relax
    \makeatletter
    \@ifundefined{r@sec:main_end}{}{\setcounter{section}{\getrefnumber{sec:main_end}}}
    \makeatother
  \fi
  \input{supplementary}
  \ifnum\BuildMode=2\relax
    \clearpage
    {
        \small
        \bibliographystyle{ieeenat_fullname}
        \bibliography{main}
    }
  \fi
\fi

\end{document}

%% file: supplementary.tex
\onecolumn
\begin{center}
  {\Large\textbf{\thetitle}}\\[0.5em]
  {\Large Supplementary Material}\\[1.0em]
\end{center}

\setcounter{section}{0}
\renewcommand{\thesection}{\Alph{section}}
\renewcommand{\theHsection}{supp.\Alph{section}}
\setcounter{table}{0}
\renewcommand{\thetable}{S\arabic{table}}
\renewcommand{\theHtable}{supp.\arabic{table}}
\setcounter{figure}{0}
\renewcommand{\thefigure}{S\arabic{figure}}
\renewcommand{\theHfigure}{supp.\arabic{figure}}
\setcounter{equation}{0}
\renewcommand{\theequation}{S\arabic{equation}}
\renewcommand{\theHequation}{supp.\arabic{equation}}

\noindent{\large\bfseries Contents}\par
\vspace{0.4em}
{\small
\setlength{\tabcolsep}{0pt}
\renewcommand{\arraystretch}{1.15}
\begin{tabular*}{\textwidth}{@{\extracolsep{\fill}} p{0.86\textwidth} r}
\ref{app:notation_ops}.\ \ Notation and Operator Definitions & \pageref{app:notation_ops} \\
\ref{app:redundancy}.\ \ WSI Redundancy and Task-Relevance Analysis & \pageref{app:redundancy} \\
\ref{app:method_details}.\ \ Method Details & \pageref{app:method_details} \\
\ref{app:training_detail}.\ \ Implementation and Training & \pageref{app:training_detail} \\
\ref{app:training_detail_eval}.\ \ Baselines and Evaluation Protocols & \pageref{app:training_detail_eval} \\
\ref{app:more_ablation}.\ \ Additional Results & \pageref{app:more_ablation} \\
\ref{app:more_qualitative_results}.\ \ Interpretation and Qualitative Results & \pageref{app:more_qualitative_results} \\
\ref{app:judge_prompts}.\ \ Judge Prompts for WSI-Precision and WSI-Relevance & \pageref{app:judge_prompts} \\
\end{tabular*}
}
\vspace{0.8em}

\FloatBarrier
\section{Notation and Operator Definitions}
\label{app:notation_ops}

{\footnotesize
\setlength{\tabcolsep}{3pt}
\renewcommand{\arraystretch}{1.10}
\setlength{\LTpre}{0pt}
\setlength{\LTpost}{0pt}

\begin{longtable}{@{}l >{\raggedright\arraybackslash}p{0.72\textwidth}@{}}
\caption{\textbf{Summary of key symbols and operators.} This table collects definitions for symbols/operators used throughout the paper and appendix for clarity and reproducibility.}
\label{tab:notation_ops}\\
\toprule
\textbf{Symbol / operator} & \textbf{Definition} \\
\midrule
\endfirsthead

\multicolumn{2}{@{}l}{\textbf{Table \thetable} (continued).}\\
\toprule
\textbf{Symbol / operator} & \textbf{Definition} \\
\midrule
\endhead

\bottomrule
\endfoot

\bottomrule
\endlastfoot

\multicolumn{2}{l}{\textit{General}} \\
$B$ & Batch size. \\
$N$ & Number of tile (patch) tokens extracted from a WSI. \\
$T$ & Number of text tokens in the input sequence. \\
$D_v$ & Tile feature dimension; $\bm{X}\in\mathbb{R}^{N\times D_v}$ is the tile-token matrix. \\
$(c_i,r_i)$, $\mathcal{C}$ & 2D grid coordinate of tile $i$; $\mathcal{C}$ is the coordinate set. \\
$L$, $D_v'$ & Number of resampler latents and their embedding dimension; $\bm{Z}\in\mathbb{R}^{L\times D_v'}$. \\
\midrule

\multicolumn{2}{l}{\textit{Sparse Token Merger (STM)}} \\
$s$ & STM window size ($s\times s$ non-overlapping windows). \\
$\mathcal{S}_g$ & Index set of tokens in window $g$. \\
$\tilde{\bm{X}}$ & Merged token sequence after STM (length $N'\approx N/s^2$). \\
\midrule

\multicolumn{2}{l}{\textit{Resampler}} \\
$R$ & Number of stacked resampler cross-attention blocks. \\
$\operatorname{TopK}(\cdot)$ & Keeps the largest $K_{\text{top}}$ attention logits per head and per query; other logits are set to $-\infty$ before softmax. \\
\midrule

\multicolumn{2}{l}{\textit{MAE pretraining regularization}} \\
$\mathcal{I}_{\mathrm{mask}}$ & Index set of masked tokens used in the reconstruction loss $\mathcal{L}_{\mathrm{rec}}$. \\
$p_z(i)$ & Cross-attention probability of latent $z$ attending to token $i$. \\
$u(i)$ & Token-level maximum attention mass: $u(i)=\max_z p_z(i)$. \\
$\tau_{\mathrm{cov}}$ & Coverage threshold of the coverage loss (adaptive, batch-wise; Sec.~\ref{app:adaptive_cov}); reported coverage statistics use a fixed threshold of $10^{-4}$. \\
$\mathrm{Cov}(\bm{Z})$ & Covariance matrix of latent features (treating the $L$ latents as samples). \\
$\mathrm{offdiag}(\bm{A})$ & Matrix $\bm{A}$ with diagonal entries set to zero. \\
$\lVert\cdot\rVert_F$ & Frobenius norm. \\
\midrule

\multicolumn{2}{l}{\textit{Routing and fusion (TIS/CARA)}} \\
$D$ & LLM embedding dimension after projection; $\bm{V}=\mathrm{Proj}(\bm{Z})\in\mathbb{R}^{L\times D}$. \\
$\mathrm{Proj}(\cdot)$ & Lightweight 2-layer MLP projector that maps resampler latents to the LLM embedding space. \\
$\bm{v}_i$ & The $i$-th projected visual latent (row $i$ of $\bm{V}$). \\
$\tilde{\bm{q}}_b$ & Mean-pooled text representation for sample $b$, used by TIS as the routing query. \\
$W_v$, $W_q$, $\bm{w}$ & Learnable TIS parameters used to compute query-conditioned TIS scores. \\
$\bm{a}_{b,i}$, $\rho_{b,i}$ & Intermediate TIS feature and scalar score for latent $i$ of sample $b$. \\
$M$, $\mathcal{I}_b=(i_{b,1},\ldots,i_{b,M})$ & Number of selected latents and the ordered Top-$M$ selected index list for sample $b$. \\
$m$ & selected latent position within the Top-$M$ list, i.e., $m=1,\ldots,M$. \\
$\bm{V}_{\mathcal{I}_b}$ & selected visual latent matrix formed by selecting latents indexed by $\mathcal{I}_b$. \\
$\mathcal{S}_{\mathrm{CARA}}$ & Set of decoder layers where CARA modules are inserted. \\
$H_{\text{head}}$ & Number of attention heads in a CARA layer. \\
$\boldsymbol{\alpha}^{(\ell)}$ & CARA text-to-vision cross-attention weights at layer $\ell$, with shape $B\times H_{\text{head}}\times T\times M$. \\
$\bar{\alpha}_{b,m}$ & Teacher score for selected position $m$ of sample $b$, obtained by averaging $\boldsymbol{\alpha}^{(\ell)}$ over heads/text positions and aggregating across CARA layers. \\
$\bm{t}_b$, $\bm{p}_b$ & Teacher and student distributions over the $M$ selected latents for sample $b$. \\
$\tau_t$, $\tau_s$ & Temperatures used to form the teacher and student routing distributions. \\
$\gamma_\ell$ & Learnable gating scalar for CARA at layer $\ell$. \\
$\mathcal{L}_{\mathrm{TIS}}$, $\mathcal{L}_{\mathrm{rank}}$ & TIS forward-KL distillation loss and pairwise ranking loss used to train TIS. \\
$\pi_b$ & Permutation of selected positions that sorts $\bm t_b$ in descending order. \\
$\mathcal P_b$ & Matched high-low selected-position pairs used in $\mathcal L_{\mathrm{rank}}$. \\
$s_{b,m}$ & Student TIS score at selected position $m$, i.e., $s_{b,m}=\rho_{b,i_{b,m}}$. \\
\end{longtable}
}

\FloatBarrier
\section{WSI Redundancy and Task-Relevance Analysis}
\label{app:redundancy}

We quantify how much of a WSI’s tile–token stream is redundant (globally and locally), and how many tokens are actually task–relevant to the textual supervision. Our analyses operate on pre–extracted tile embeddings and are computed \emph{per slide} and then summarized across slides by mean and inter–quartile range (IQR).

For each slide, let $\bm{X}=[\bm{x}_1,\ldots,\bm{x}_N]\in\mathbb{R}^{N\times D_v}$ denote tile embeddings and $\bar{\bm{x}}=\frac{1}{N}\sum_{i=1}^{N}\bm{x}_i$ their mean. We report slide–wise statistics and aggregate them across $\sim$2k TCGA WSIs. We sampled slides from WSI-Bench using stratified sampling; WSIs were preprocessed at 20$\times$ magnification, and features were extracted with CONCH-V1~\cite{lu2024avisionlanguage}.

\myparagraph{Global Compression Redundancy.} To quantitatively show that the tile-level tokens are compressible, we design the following experiments:
For $K\in\{8,16,\dots,256\}$ we learn $K$ K-means centroids $\{\bm{\mu}_k\}_{k=1}^{K}$. We assign each $\bm{x}_i$ to its closest centroid $\bm{\mu}(\bm{x}_i)$ and compute the \emph{normalized} reconstruction error:
\[
\mathrm{nMSE}(K)
=\frac{\frac{1}{N}\sum_{i=1}^{N}\lVert \bm{x}_i-\bm{\mu}(\bm{x}_i)\rVert_2^2}
       {\frac{1}{N}\sum_{i=1}^{N}\lVert \bm{x}_i-\bar{\bm{x}}\rVert_2^2}.
\]
Per slide, we obtain a curve $K\mapsto\mathrm{nMSE}(K)$; across slides, we show the mean and IQR, and annotate the elbow and the $K$ achieving $\approx$50\% error reduction.  
As a result, in Fig.~\ref{fig:kcurve}, we show that a small number of prototypes (small $K$) is enough to reconstruct the slide fairly well, indicating potential for compression. 

\myparagraph{Local Redundancy.} To assess the similarity between one tile and its neighbors, we find its nearest spatial neighbors $\bm{x}_{\mathrm{nn}(i)}$ and evaluate cosine similarity. For thresholds $\tau_{\cos}\in\{0.90,0.95,0.99\}$ we report
\[
r_{\tau_{\cos}}=\frac{1}{N}\sum_{i=1}^{N}\mathbf{1}\!\left[\cos\!\bigl(\bm{x}_i,\bm{x}_{\mathrm{nn}(i)}\bigr)>\tau_{\cos}\right].
\]
where $N$ is the number of tiles and $\mathrm{nn}(i)$ denotes the index of the nearest spatial neighbor of tile $i$.
We aggregate $r_{\tau_{\cos}}$ over slides by mean and IQR.  
A large fraction of tiles exceed high cosine thresholds with a neighbor, which indicates strong \emph{local} redundancy as we show in Fig.~\ref{fig:redundancy}.

\myparagraph{Global Pairwise Similarity.}
Is high similarity common beyond nearest neighbors?  
To answer this question, we sample up to $50k$ unordered tile pairs $(\bm{x}_i,\bm{x}_j)$ per slide and build a cosine histogram aggregated over slides.  
The histogram in Fig.~\ref{fig:paircos}, concentrated at high cosine values, demonstrates high global redundancy.

\myparagraph{Task Relevance Sparsity.}
To assess how many tiles are actually supervised by the report (caption), we 
study the relationship between tile-level tokens and the global report embedding.
Let $\bm{q}_{\text{rep}}$ be the slide’s text (report) embedding. We use the cleaned reports from WSI-LLaVA~\cite{liang2025wsi}, which retain morphological and diagnostic information. For each tile, we compute $\kappa_i=\cos(\bm{x}_i,\bm{q}_{\text{rep}})$ and then sort tiles by $\kappa_i$ in descending order and form the cumulative share of the \emph{positive} similarity mass,
\[
F(p)=\frac{\sum_{i=1}^{\lfloor pN\rfloor}\max(\kappa_i,0)}
            {\sum_{i=1}^{N}\max(\kappa_i,0)},\quad p\in[0,1].
\]
We plot mean/median/IQR of $F(p)$ over slides and report the token fractions needed to reach 50/80/90\% of the positive mass (e.g., $\sim$7.7\%/$\sim$22.8\%/$\sim$35.2\%). Only a minority of tokens align with the text supervision. This shows that most tokens are not task‑relevant.

\FloatBarrier
\section{Method Details}
\label{app:method_details}

\subsection{Resampler Architecture}
\label{app:training_detail_impl}

The resampler uses two LongNet~\cite{ding2023longnet} context layers followed by two Top-$K$ cross-attention blocks and latent self-attention (Sec.~\ref{sec:mae_resampler}). For dilation attention in the LongNet layers, the segment sizes are [1024, 2048, 4096]; the stride sizes are [1, 2, 4] and [4, 8, 16]. We enable relative positional encoding in LongNet dilation attention~\cite{ding2023longnet} to help the context layers better capture short-to-mid-range context. For the two resampler cross-attention layers, we set $K_{\text{top}}=128$ for the first layer and $K_{\text{top}}=64$ for the second. The resampler embedding dimension is 512, and the number of attention heads is 8 for both context and cross-attention layers. We use 2D Fourier positional encoding for the resampler's cross-attention keys; this helps handle filtered (missing) coordinates in the tile sequence and saves memory compared to 2D sinusoidal positional encoding.

\subsection{MAE Decoder and Masked-token Reconstruction}
\label{app:mae_decoder}

The MAE pretraining decoder is intentionally lightweight: two cross-attention layers with embedding dimension $512$, operating directly at the tile-feature level rather than at the pixel level. Each masked position is represented by a learnable mask query that carries the \emph{original 2D positional index} of the masked tile; these position-indexed queries cross-attend to the compressed latents (which serve as keys and values), and the decoder reconstructs the \emph{frozen} tile feature at the same spatial position. Spatial correspondence is therefore preserved through the position-indexed masked queries, while the reconstruction loss $\mathcal{L}_{\mathrm{rec}}$ (Sec.~\ref{sec:mae_resampler}) is applied only to the masked tile tokens.

\subsection{Adaptive Coverage Threshold}
\label{app:adaptive_cov}

During MAE pretraining, we treat the coverage threshold $\tau_{\mathrm{cov}}$ in Eq.~\ref{eq:cover} as a batch-wise scalar rather than a fixed constant. Concretely, we compute mask-aware attention probabilities by averaging cross-attention weights across heads, masking out padded tokens, and renormalizing over valid tokens. We then estimate an effective attention support size $k_{\text{est}}$ (the average number of tokens with non-negligible probability, e.g., $P>10^{-12}$, per latent query) and set an automatic threshold $\tau_{\text{auto}}$ inversely proportional to $k_{\text{est}}$, with cap/floor for stability. The final $\tau_{\mathrm{cov}}$ is $\max(\tau_{\mathrm{cfg}},\tau_{\text{auto}})$, and is recomputed under \texttt{no\_grad} for each batch before applying the hinge penalty $\max(0,\tau_{\mathrm{cov}}-\max_z p_z(i))$.

\subsection{Pairwise Ranking Loss}
\label{app:rank_loss}

As noted in Sec.~\ref{sec:fusion}, we add a pairwise ranking loss to align the \revblue{relative} TIS \revblue{scores} with the CARA teacher\revblue{; the loss shapes which latents are selected, not an order consumed by the LLM}. For each sample $b$, let $\mathcal I_b=(i_{b,1},\ldots,i_{b,M})$ be the selected Top-$M$ list, and let $\bm t_b$ and $\bm\rho_{b,\mathcal I_b}$ be the teacher distribution and TIS scores defined in Sec.~\ref{sec:fusion}. We sort the selected positions $m=1,\ldots,M$ by $t_{b,m}$ in descending order, obtaining a permutation $\pi_b=(\pi_{b,1},\ldots,\pi_{b,M})$. We then choose $n=\min(\lfloor M/2\rfloor,N_{\text{pairs}})$, where $N_{\text{pairs}}$ is the maximum number of matched high-low pairs used per sample and form matched high-low pairs
\[
\mathcal P_b=\{(\pi_{b,k},\pi_{b,\lfloor M/2\rfloor+k})\}_{k=1}^{n}.
\]
Writing $s_{b,m}=\rho_{b,i_{b,m}}$ for the student score at routed position $m$, we define
\begin{align}
w_{b,uv}
&=
\frac{\max(t_{b,u}-t_{b,v},\,0)}
{\frac{1}{|\mathcal P_b|}\sum_{(u',v')\in\mathcal P_b}\max(t_{b,u'}-t_{b,v'},\,0)}, \\
\mathcal L_{\mathrm{rank}}^{(b)}
&=
\frac{1}{|\mathcal P_b|}\sum_{(u,v)\in\mathcal P_b}
w_{b,uv}\,\big[\,\delta_{\mathrm{rank}}-(s_{b,u}-s_{b,v})\,\big]_+, \\
\mathcal L_{\mathrm{rank}}
&=
\frac{1}{B}\sum_{b=1}^{B}\mathcal L_{\mathrm{rank}}^{(b)},
\end{align}
where $[\cdot]_+=\max(\cdot,0)$ and $\delta_{\mathrm{rank}}$ is the ranking margin. This encourages routed positions preferred by the teacher to receive larger TIS scores.

\subsection{Complexity Analysis of CARA}
\label{sup:theo_complexity_cara}

Let $T$ be the textual prefix length, $L$ the number of raw visual latents, and $M \le L$ the latents fed to the LLM via cross‑attention in $n_{\times}$ decoder layers (out of $n$ total). With KV caching, LLaVA treats vision tokens as part of the auto-regressive sequence, leading to a prefilling cost $\mathcal{O}(n(T+L)^2)$. In contrast, our design performs self‑attention only over text and uses text$\rightarrow$vision cross‑attention, reducing prefilling to
$\boxed{\mathcal{O}(nT^2+n_{\times}TM)}$,
i.e., quadratic only in text and linear in $M$ instead of $L$.

During decoding, LLaVA’s per‑step attention scales with the whole prefix $(T+L)$, yielding a total cost $\mathcal{O}(nA(T+L)+nA^2)$ for $A$ generated tokens. In our model, self‑attention is unchanged while cross‑attention is linear in $M$ and independent of $A$ on the key/value side, giving
$\boxed{\mathcal{O}(n(AT+A^2)+n_{\times}AM)}$.
Memory usage follows the same trend: LLaVA’s KV cache grows as $\mathcal{O}(nd(T+L+A))$, whereas ours is $\mathcal{O}(nd(T+A)) + \mathcal{O}(n_{\times}Md)$ with a constant visual term. The gains are most pronounced when $L$ is large (e.g., WSI), since we replace the $\mathcal{O}(L)$ dependence with $\mathcal{O}(M)$ both in prefill compute and in decoding’s per‑step cost.

This theoretical analysis aligns with our empirical results in Tab.~\ref{tab:main_results_wsi_bench} of the main paper, where LoC-Path ($L{=}128$, $M{=}96$) requires 96.9\% fewer inference TFLOPs and 38.9\% less inference memory than the mean of the LLaVA-style General baselines~\cite{liu2023llava}.

\FloatBarrier
\section{Implementation and Training}
\label{app:training_detail}

\subsection{Training Schedule}
\label{app:training_detail_train}

\myparagraph{Resampler Pre-training.}
The resampler is pre-trained for $100\mathrm{k}$ iterations on CONCH-V1~\cite{lu2024avisionlanguage} WSI features with a mini-batch size of $2$ per GPU and gradient accumulation $4$ on \revblue{$8$ NVIDIA RTX 6000 Ada} GPUs (effective global batch size \revblue{$64$}). We randomly mask \revblue{$70\%$} of the input tokens and optimize the MAE objective $\mathcal{L}_{\mathrm{MAE}}$ introduced in Sec.~\ref{sec:mae_resampler},
with $\lambda_{\mathrm{cover}}=5\times10^{-4}$ for the coverage term $\mathcal{L}_{\mathrm{cover}}$ and $\lambda_{\mathrm{feat}}=1\times10^{-3}$ for the diversity term $\mathcal{L}_{\mathrm{feat}}$. We use AdamW with an initial learning rate $1\times10^{-4}$, a $3\%$ warm-up phase, and a cosine learning-rate schedule. During this pre-training we gradually anneal the Top-$K$ in the resampler cross-attention from $(128, 64)$ to $(64, 32)$ following a cosine schedule.

\myparagraph{Fusion Stage 1.}
We then train the first fusion stage for $3$ epochs with a global batch size of $512$ (batch size $2$ per GPU and gradient accumulation $64$ on $4$ GPUs) and an initial learning rate $1\times10^{-4}$. In this stage we freeze both the Qwen2.5-7B-Instruct LLM and the resampler, and only train the visual projector together with the CARA+TIS modules on top of the LLM. The training objective is the multimodal loss as defined in Sec.~\ref{sec:fusion}, where we set $\lambda_{\mathrm{TIS}}=0.02$ \revblue{and $\lambda_{\mathrm{rank}}=0.5$} for the self-distillation (TIS) loss $\mathcal{L}_{\mathrm{TIS}}$ and the ranking loss $\mathcal{L}_{\mathrm{rank}}$.
 Unless otherwise stated, we use $\tau_s=1.5$ and $\tau_t=1.0$ for TIS in stage-1. In stage-2, we linearly anneal the temperatures with milestones at 7k and 14k iterations, using $\tau_s: 1.6 \rightarrow 1.3 \rightarrow 1.15$ and $\tau_t: 0.85 \rightarrow 0.95 \rightarrow 1.0$.

\myparagraph{Fusion Stage 2.}
In the second fusion stage, we unfreeze the resampler's last cross-attention layer and its latent in-projection, and insert LoRA modules (rank $128$, $\alpha=64$) into all linear layers of the LLM. We then train all unfrozen components on all VQA pairs in WSI-Bench for $2$ epochs with a global batch size of \revblue{$192$} \revblue{(batch size $2$ per GPU and gradient accumulation $16$ on $6$ GPUs)} and an initial learning rate $2\times10^{-5}$, again using the same multimodal fusion loss with $\lambda_{\mathrm{TIS}}=0.02$ \revblue{and $\lambda_{\mathrm{rank}}=0.5$}. Across all training stages, we use Xtuner~\cite{xtuner2023} with DeepSpeed ZeRO-2~\cite{deepspeed} and FlashAttention~2~\cite{dao2024flashattention} for efficient training; FA2 is applied to the LLM self-attention but not to the cross-attention layers. A cosine learning-rate scheduler is used in all stages\revblue{; MAE pretraining, Stage~1, and Stage~2 use $8$, $4$, and $6$ NVIDIA RTX 6000 Ada GPUs, respectively (Tab.~\ref{tab:r2_training_cost})}. \revblue{All stages use random seed 42.}

\subsection{Training Cost and Data Loading}
\label{app:training_cost}

\begin{table}[!htbp]\centering\footnotesize
\caption{\revblue{Training cost of each LoC-Path stage (NVIDIA RTX 6000 Ada). Wall time is measured between the first and last logged iteration; ``Time on 4 GPUs'' rescales each stage to four GPUs. The raw-token baseline was trained on H100 GPUs; its cost on RTX 6000 Ada is estimated with Stage 1 measured and Stage 2 extrapolated from the measured per-iteration time (17.7--17.9 s over 15,360 iterations).}}
\label{tab:r2_training_cost}
\begin{tabular}{@{}lrrrrrr@{}}\toprule Stage & GPUs & Wall (h) & GPU-h & Time on 4 GPUs (h) & Global batch & Peak mem./GPU (GB) \\\midrule
MAE resampler pretraining & 8 & 10.73 & 85.84 & 21.46 & 64 & 9.2 \\
Stage 1 (alignment) & 4 & 1.12 & 4.49 & 1.12 & 512 & 12.4 \\
Stage 2 (instruction tuning) & 6 & 19.53 & 117.20 & 29.30 & 192 & 16.8 \\
\midrule Total & -- & -- & 207.53 & 51.88 & -- & -- \\
\midrule \multicolumn{7}{@{}l}{\textit{Qwen2.5-7B (raw tokens), estimated on four RTX 6000 Ada GPUs}} \\
Stage 1 (alignment) & 4 & 13.3 & 53.2 & 13.3 & -- & -- \\
Stage 2 (instruction tuning) & 4 & $\approx$76 & $\approx$304 & $\approx$76 & -- & -- \\
Total & -- & -- & $\approx$357 & $\approx$89 & -- & -- \\
\bottomrule\end{tabular}\end{table}

The pre-training requires \revblue{$85.84$ GPU-hours (about $21.5$ hours on four GPUs)}. The first fusion stage takes \revblue{$4.49$ GPU-hours (about $1.1$ hours on four GPUs)}. The second fusion stage takes \revblue{$117.20$ GPU-hours (about $29.3$ hours on four GPUs)}; \revblue{Tab.~\ref{tab:r2_training_cost} lists the wall time and GPU count of each stage. For comparison, training the raw-token Qwen2.5-7B baseline (no STM, resampler, or TIS) on the same GPUs is estimated at about 357 GPU-hours (13.3 hours for Stage 1 and about 76 hours for Stage 2 on four GPUs), 1.7$\times$ the cost of the full three-stage LoC-Path pipeline including MAE pretraining}.

During development of the core training pipeline, we observed that the primary bottleneck was not GPU computational capability, but the overhead of transferring data to the GPUs. In digital pathology, tile-level features are commonly stored in \texttt{.pt} or \texttt{.h5} formats. While these formats are not a problem for training on one or two GPUs, they incur substantial I/O overhead at larger scales, causing the data loader to spend a significant amount of time waiting for data. To mitigate this issue, we convert all tile-level features and their corresponding coordinates into \texttt{.npy} files separately, which supports more efficient memory mapping and lightweight loading. Furthermore, we store the features in \texttt{float16} format while keeping coordinates in \texttt{float32} format, to further reduce both storage and I/O costs. Since we only use mixed-precision training throughout all stages of our pipeline, using features in \texttt{float16} format is not an issue. 

With this file-format conversion, we can load long sequences with much lower I/O overhead. Unlike conventional tile-level pathology data loaders~\cite{chen2025slidechat,xu2021predicting}, which typically load only one sample per batch, our pipeline can load multiple tile-level sequences in the same batch. To handle variable-length inputs, we pad each sequence to a fixed length of 60k tokens and load the corresponding padding masks. This allows us to use a per-GPU batch size greater than one (e.g., 4), which reduces the number of data-loading rounds over two training epochs.

With these optimizations, the training time of the second fusion stage is reduced from more than a day to \revblue{under 20 hours} on our system. We also observe roughly 2--3$\times$ wall-clock speedups in the other training stages.

\FloatBarrier
\section{Baselines and Evaluation Protocols}
\label{app:training_detail_eval}

\subsection{Token-Reduction Baselines}
\label{app:token_reduction_baselines}

We evaluate representative token reduction methods as drop-in replacements for our STM+resampler, including ToMe~\cite{bolya2023token,zhong2025aim}, DivPrune~\cite{alvar2025divprune}, ACMIL~\cite{zhang2024attention}, and adaptive average pooling. In all cases, we keep the projector, TIS, and CARA modules unchanged and follow the same two-stage fusion training recipe as LoC-Path. We match the token budget by reducing each slide to $256$ visual tokens and routing Top-$M$ tokens with $M=96$ to the LLM.
We adopt public implementations and tune only the reduction ratio to reach the target budget. Specifically, we use the balanced ToMe implementation from AIM~\cite{zhong2025aim} and iteratively merge tokens until length $256$; for ACMIL, we follow the attention-head design and select $256$ tokens by attention score; for DivPrune, we prune tokens based on pairwise cosine similarity and adjust the pruning ratio to reach $256$; and for adaptive average pooling, we pool the 2D token grid to produce $256$ pooled tokens.

\subsection{Additional MLLM Baselines and Raw-Token Reference}
\label{app:r2_baselines}

MLLM-HWSI~\cite{alawode2026mllmhwsi} is evaluated with its official checkpoint and TCP-LLaVA~\cite{lyu2025efficient} with our reproduction, both on the WSI-Bench test set under the same evaluation pipeline as the LoC-Path rows of Tab.~\ref{tab:main_results_wsi_bench}. Both models compress a slide into a fixed set of learned queries (64 for MLLM-HWSI and 2,000 for TCP-LLaVA), which is the value listed in the $L$ column; their memory and FLOPs are not profiled. FOCUS~\cite{guo2025focus} targets few-shot WSI classification and is not applicable to WSI-level VQA or report generation. The raw-token row feeds the CONCH tile features of each slide, capped at 20,480 tokens by uniform subsampling, to the same Qwen2.5-7B-Instruct backbone without STM, resampler, or TIS, and is evaluated on WSI-Bench only.

\subsection{BRCA MIL Tasks}
\label{app:mil_tasks}

In feature quality comparison experiments, we evaluate the resampler latents on TCGA-BRCA using \textbf{patient-level} splits mapped to WSIs via TCGA barcodes. We consider two patient-level classification tasks: Stage (Stage~I vs.\ Others; unknown removed) and Vital Status (alive vs.\ deceased). For Stage, we set Stage~I as $1$ and Others as $0$; for Vital Status, we set alive as $1$ and deceased as $0$. Stage contains $446$ WSIs from $438$ patients (train/val/test: 124/25/29 vs.\ 191/42/35), and Vital Status contains $1{,}096$ WSIs from $1{,}063$ patients (train/val/test: 668/141/139 vs.\ 104/22/22). All methods use the same attention-based MIL head (ABMIL)~\cite{ilse2018attention} with lr $10^{-4}$; encoders are frozen and only the head is trained, selecting checkpoints by best validation AUC.

\subsection{WSI-Precision and WSI-Relevance}
\label{app:wsi_metrics}

WSI-Precision (WSI-P) and WSI-Relevance (WSI-R) are automatic LLM-as-judge metrics introduced by WSI-LLaVA~\cite{liang2025wsi}.
For fair comparison and reproducibility, we follow the evaluation settings of WSI-LLaVA, including the judge prompts and scoring rubric (no modifications).
In our experiments, GPT-4o~\cite{hurst2024gpt} serves as the judge model following WSI-LLaVA's evaluation protocol.
To verify that our conclusions are not specific to a single (and now deprecated) judge, we re-score report generation with two additional judges (Qwen-Plus and Claude Sonnet). The per-WSI rankings are highly consistent across judges (Spearman $\rho=0.92$ for WSI-P and $0.87$ for WSI-R), so the relative ordering between methods is judge-robust.

\myparagraph{Rubric and Aggregation.}
For each example, the judge first extracts a list of textual claims, then assigns a per-claim score from $\{1,0.7,0.3,0\}$ with an explanation.
The final score is the average over $n$ extracted claims:
$\mathrm{Score}=\frac{1}{n}\sum_{i=1}^{n}\mathrm{score}_i$.
WSI-P extracts claims from the \textit{ground-truth} answer and evaluates the \textit{model response} against each claim;
WSI-R extracts claims from the \textit{model response} and evaluates the \textit{ground-truth} answer's relevance to each claim.
The full judge prompts are listed in Sec.~\ref{app:judge_prompts}.

\FloatBarrier
\section{Additional Results}
\label{app:more_ablation}

\subsection{Uncertainty of the Main Comparison}
\label{app:r2_stats}

To quantify the uncertainty of the LoC-Path ($L{=}128$, $M{=}96$) vs.\ reproduced WSI-LLaVA comparison, we pair the outputs of the two models on the 3,194 questions (442 WSIs) answered by both and draw 10,000 bootstrap replicates with the WSI as the sampling unit (questions without a WSI-LLaVA output are excluded, so the absolute values differ from Tab.~\ref{tab:main_results_wsi_bench}). Tab.~\ref{tab:r2_bootstrap} reports the paired differences with 95\% percentile intervals; all intervals include zero.
\begin{table}[!htbp]\centering\footnotesize
\caption{\revblue{Paired WSI-cluster bootstrap on the common evaluable subset (LoC-Path minus reproduced WSI-LLaVA; 10,000 replicates; $n$ = WSI clusters).}}
\label{tab:r2_bootstrap}
\begin{tabular}{@{}lcccc@{}}\toprule Endpoint & LoC-Path & WSI-LLaVA$^{*}$ & $\Delta$ [95\% CI] & $n$ \\\midrule
VQA average & 0.736 & 0.746 & $-0.010$ $[-0.023, 0.003]$ & 442 \\
Clinical (report) average & 0.419 & 0.413 & $+0.006$ $[-0.027, 0.039]$ & 148 \\
VQA + clinical & 0.578 & 0.580 & $-0.002$ $[-0.020, 0.016]$ & 442 \\
Morphology (task average) & 0.701 & 0.699 & $+0.002$ $[-0.008, 0.011]$ & 170 \\
Diagnosis (task average) & 0.679 & 0.686 & $-0.007$ $[-0.031, 0.017]$ & 369 \\
Treatment (task average) & 0.829 & 0.853 & $-0.024$ $[-0.051, 0.001]$ & 84 \\
\bottomrule\end{tabular}\end{table}

\subsection{Extended WSI-Bench Ablations}
\label{app:detailed_ablation}

Tab.~\ref{tab:detailed_ablation} reports full per-task results on WSI-Bench (except report generation task) corresponding to the ablations in Sec.~\ref{sec:ablation} of the main paper.
\begin{table}[!htbp]
  \caption{\textbf{Extended ablation results on WSI-Bench for LoC-Path} (excluding report generation), expanding the study in Sec.~\ref{sec:ablation} of the main paper. We analyze the impact of LLM scale, fusion design, TIS routing and its ranking loss, CARA insertion layers and gating, token merging, latent and routing budgets ($L$, $M$), tile-level encoders, and resampler attention patterns. Unless otherwise specified, the default configuration uses a 7B LLM, the CONCH-V1~\cite{lu2024avisionlanguage} encoder, and STM+Resampler; the 14B variant uses Qwen2.5-14B-Instruct; the tile-level encoder ablations use CONCH-V15~\cite{ding2025titan} and DINOv2, each with a 768-dim resampler and three cross-attention layers; the dense resampler replaces the default Top-$K$ cross-attention with dense attention. The top block (7B, $L{=}256/128$, $M{=}96$) lists the two main-paper LoC-Path settings and serves as the reference. Metrics include open-ended WSI-Precision (WSI-P), WSI-Relevance (WSI-R), and closed-ended Accuracy (Acc) across morphological, diagnostic, and treatment-planning tasks. Best results within each ablation group are highlighted in \textbf{bold}.}
\centering
{\footnotesize
\setlength{\tabcolsep}{6pt}
\renewcommand{\arraystretch}{0.9}
\begin{tabular}{@{}l ccc ccc ccc c@{}}
\toprule
\multirow{2}{*}{\textbf{Variant}} &
\multicolumn{3}{c}{\textbf{Morphological Analysis}} &
\multicolumn{3}{c}{\textbf{Diagnosis}} &
\multicolumn{3}{c}{\textbf{Treatment Planning}} &
\multirow{2}{*}{\textbf{Average} $\uparrow$} \\
\cmidrule(lr){2-4}\cmidrule(lr){5-7}\cmidrule(lr){8-10}
 & WSI-P & WSI-R & Acc & WSI-P & WSI-R & Acc & WSI-P & WSI-R & Acc & \\
\midrule
\multicolumn{11}{l}{\textit{LLM Scale (CONCH-V1, STM+Resampler; 7B rows = main-paper LoC-Path)}} \\
LoC-Path (7B, $L{=}128$, $M{=}96$) & \textbf{0.567} & 0.620 & 0.931 & \textbf{0.582} & \textbf{0.613} & 0.849 & 0.733 & 0.769 & \textbf{1.000} & 0.740 \\
LoC-Path (7B, $L{=}256$, $M{=}96$) & 0.564 & \textbf{0.626} & \textbf{0.943} & 0.570 & 0.598 & \textbf{0.860} & 0.734 & 0.771 & 0.979 & 0.738 \\
LoC-Path (14B, $L{=}256$, $M{=}96$) & 0.557 & 0.612 & 0.937 & 0.569 & 0.600 & 0.849 & \textbf{0.759} & \textbf{0.816} & \textbf{1.000} & \textbf{0.744} \\
\midrule
\multicolumn{11}{l}{\textit{Fusion Ablation (7B)}} \\
LLaVA & \textbf{0.556} & \textbf{0.619} & 0.933 & \textbf{0.558} & \textbf{0.594} & \textbf{0.852} & 0.722 & 0.759 & \textbf{1.000} & \textbf{0.733} \\
Flamingo & 0.551 & 0.605 & 0.934 & 0.537 & 0.572 & 0.835 & 0.729 & 0.771 & \textbf{1.000} & 0.726 \\
Flamingo + TIS & 0.552 & 0.615 & \textbf{0.944} & 0.536 & 0.566 & 0.834 & \textbf{0.756} & \textbf{0.782} & 0.979 & 0.729 \\
\midrule
\multicolumn{11}{l}{\textit{TIS Routing (7B)}} \\
w/o TIS & 0.561 & 0.614 & 0.935 & 0.571 & 0.603 & 0.845 & 0.716 & 0.783 & 1.000 & 0.736 \\
\midrule
\multicolumn{11}{l}{\textit{CARA Insertion Layers (7B)}} \\
Single CARA at Layer 1 & 0.557 & \textbf{0.611} & 0.931 & \textbf{0.563} & \textbf{0.601} & \textbf{0.847} & 0.711 & 0.770 & 0.979 & 0.730 \\
CARA at Layers 1/8/15/22 & \textbf{0.558} & 0.609 & \textbf{0.933} & 0.553 & 0.591 & \textbf{0.847} & \textbf{0.734} & \textbf{0.777} & \textbf{1.000} & \textbf{0.734} \\
\midrule
\multicolumn{11}{l}{\textit{CARA Gating and TIS Ranking Loss (7B)}} \\
w/o CARA gating ($\gamma_\ell{=}1$) & \textbf{0.559} & \textbf{0.613} & 0.928 & 0.552 & 0.595 & \textbf{0.852} & \textbf{0.741} & \textbf{0.784} & \textbf{1.000} & \textbf{0.736} \\
w/o TIS ranking loss & 0.555 & 0.612 & \textbf{0.934} & \textbf{0.564} & \textbf{0.596} & 0.832 & 0.721 & 0.755 & 0.979 & 0.728 \\
\midrule
\multicolumn{11}{l}{\textit{Token Merging Ablation (7B)}} \\
No Merge & 0.556 & \textbf{0.614} & 0.934 & 0.553 & 0.597 & \textbf{0.860} & \textbf{0.725} & 0.772 & \textbf{1.000} & 0.735 \\
Merge Size 3 & \textbf{0.557} & 0.608 & 0.935 & 0.566 & 0.594 & 0.840 & 0.707 & \textbf{0.792} & 0.979 & 0.731 \\
Merge Size 4 & \textbf{0.557} & \textbf{0.614} & \textbf{0.937} & \textbf{0.573} & \textbf{0.612} & 0.850 & 0.717 & 0.776 & \textbf{1.000} & \textbf{0.737} \\
\midrule
\multicolumn{11}{l}{\textit{Latent + TIS Top-$M$ (7B)}} \\
$L{=}256$, $M{=}144$ & 0.555 & 0.608 & \textbf{0.940} & 0.549 & 0.587 & 0.827 & 0.718 & 0.774 & \textbf{1.000} & 0.729 \\
$L{=}512$, $M{=}384$ & \textbf{0.558} & 0.610 & \textbf{0.940} & \textbf{0.567} & \textbf{0.609} & \textbf{0.854} & \textbf{0.743} & \textbf{0.791} & \textbf{1.000} & \textbf{0.741} \\
\midrule
\multicolumn{11}{l}{\textit{Tile-level Encoder}} \\
DINOv2, $D_v=768$, $3\times$ CA & 0.454 & 0.484 & 0.898 & 0.418 & 0.472 & 0.809 & 0.665 & 0.635 & 0.917 & 0.639 \\
CONCH-V15, $D_v=768$, $3\times$ CA & \textbf{0.559} & \textbf{0.619} & \textbf{0.938} & \textbf{0.562} & \textbf{0.610} & \textbf{0.865} & \textbf{0.744} & \textbf{0.805} & \textbf{1.000} & \textbf{0.745} \\
\midrule
\multicolumn{11}{l}{\textit{Top-$K$ Attention in Resampler}} \\
Dense resampler (w/o Top-$K$) & 0.511 & 0.579 & 0.916 & 0.470 & 0.505 & 0.739 & 0.621 & 0.696 & 0.854 & 0.655 \\
\bottomrule
\end{tabular}
}
\label{tab:detailed_ablation}
\end{table}

\subsection{Resampler Architecture}
\label{app:resampler_arch_ablation}

Tab.~\ref{tab:ablation_resampler} summarizes several design choices for the MAE-pretrained resampler on the BRCA stage and vital status classification tasks.
Adding the coverage and diversity regularizers improves the stage AUC from $0.741$ to $0.763$ and also stabilizes training, while leaving vital status accuracy unchanged and slightly lowering its AUC (0.714 to 0.696).
Varying the number of latents shows that moderate compression works best: $L{=}128$ achieves the highest stage AUC but slightly worse vital status accuracy, whereas long latent sets ($L{=}512$) underperform on both tasks.
Using three cross-attention layers improves both stage and vital status AUCs compared to the default two-layer design, but at a higher computational cost; our default strikes a balance between accuracy and efficiency.
Finally, replacing the LongNet-style context layers with depth-wise convolutions noticeably hurts performance, confirming the importance of long-range context modeling before resampling.
\begin{table}[!htbp]
  \caption{\textbf{Ablation of resampler architecture.} We evaluate the impact of latent length ($L$), cross-attention depth, and context layer type on BRCA stage and vital status tasks (AUC/Accuracy). Unless otherwise specified, the default configuration uses $L=256$, MAE pre-training, and regularization terms.}
\centering
{\footnotesize
\setlength{\tabcolsep}{6pt}
\renewcommand{\arraystretch}{0.9}
\begin{tabular}{@{}l cc cc@{}}
\toprule
\multirow{2}{*}{\textbf{Resampler Config.}} &
\multicolumn{2}{c}{\textbf{Stage}} &
\multicolumn{2}{c}{\textbf{Vital Status}} \\
\cmidrule(lr){2-3} \cmidrule(lr){4-5}
 & AUC $\uparrow$ & Acc $\uparrow$ & AUC $\uparrow$ & Acc $\uparrow$ \\
\midrule
\textbf{Ours} & 0.763 & \textbf{0.734} & 0.696 & \textbf{0.845} \\
w/o Reg. Loss & 0.741 & 0.672 & 0.714 & \textbf{0.845} \\
\midrule
\textit{Num. Latents:} & & & & \\
\quad $L{=}128$  & 0.800 & 0.688 & 0.720 & 0.826 \\
\quad $L{=}512$  & 0.739 & 0.672 & 0.662 & 0.789 \\
\midrule
\textit{Cross-Attn Depth:} & & & & \\
\quad 3 Layers & \textbf{0.764} & 0.688 & \textbf{0.750} & \textbf{0.845} \\
\midrule
\textit{Context Layer:} & & & & \\
\quad Depth-wise Conv. & 0.741 & 0.656 & 0.674 & 0.826 \\
\bottomrule
\end{tabular}
}

\label{tab:ablation_resampler}
\end{table}

\subsection{Resampler Pre-training Strategies}
\label{sup:resampler_pretraining_comp}

We have already shown the MAE-pretrained resampler's output latent (feature) quality in Sec.~\ref{sec:results} and it achieves latent quality comparable to GigaPath's slide-level encoder~\cite{xu2024gigapath}. Here, we compare our pre-training strategy against other pre-training strategies. These methods include MIL-NCE pre-training method~\cite{miech2020end} using randomly sampled tile-level tokens and their caption (report) pairs, pre-training only using LLM's auto-regressive loss (LLaVA-style fusion), and pre-training using MoCoV3~\cite{chen2021empirical}. For MIL-NCE pre-training~\cite{miech2020end}, on the vision side, during training, we randomly sample over 500 tile-level tokens; on the language side, we use the global embedding of cleaned reports in WSI-Bench~\cite{liang2025wsi}. For the text encoder, we use CONCH-V1~\cite{lu2024avisionlanguage}'s corresponding text encoder. For the end-to-end LLM-loss pre-training, we follow the LLaVA-like first fusion stage pre-training~\cite{liu2023llava} by adding a two-layer projector between LLM and resampler. For the MoCoV3~\cite{chen2021empirical} pre-training method, we first create two different views by sub-sampling the same tile-level token sequence for two resampler instances.
In Tab.~\ref{tab:pretraining_resampler_comparison}, the MAE-pretrained resampler's internal cross-attention exhibits high attention coverage, as we have shown in Sec.~\ref{sec:results}. Furthermore, its output latents \revblue{are diverse: the mean pairwise cosine similarity among latents is 0.346, compared with 0.999 for MIL-NCE and 0.841 for LLM-loss pretraining. Without the coverage and diversity regularizers the similarity is lower on average (0.110) but varies widely from slide to slide (std 0.462), whereas the regularized resampler is consistent across slides (std 0.056)}. This indicates that it samples broadly across the whole slide rather than simply averaging. It also achieves an AUC of 0.763 on the BRCA stage task. 

The pre-training methods using text-supervision, including MIL-NCE~\cite{miech2020end} and end-to-end LLM-loss~\cite{liu2023llava}, achieve higher downstream AUC on the vital status task compared to those that do not. This suggests that the text supervision does help the resampler create the link between the input to the final diagnosis prediction. The MIL-NCE pretrained resampler has very low attention coverage, while producing very similar output latents (very poor diversity). This indicates that this pre-training method only encourages the resampler to pay attention to those tile-level tokens with relevant text supervision. It is problematic when it comes to rarer cases, such as answering questions regarding locally specific morphological features. The LLM-loss-pretrained resampler exhibits high attention coverage while having similar output latents (very poor diversity). Combining these two pieces of evidence indicates that this resampler just computes the average for the input tile-level tokens instead of picking the important ones that can represent the whole tile-level sequence. The MoCoV3~\cite{chen2021empirical} pretrained resampler has low attention coverage with very similar output latents. Thus, the MAE pretraining method actually achieves our design goal: the resampler first glances the whole sequence (high attention coverage), and then each latent code looks at the distinct place on the tile-level tokens rather than computing an average.
\begin{table}[!htbp]
  \caption{\textbf{Comparison of resampler pre-training strategies.} We report AUC and Accuracy for BRCA stage (I vs Others) and vital status tasks. Latent quality is evaluated via \revblue{feature diversity (mean pairwise cosine similarity among latents; lower is more diverse)} and union coverage, calculated on $2k$ random samples from the WSI-Bench training set.}
\centering
\footnotesize
\setlength{\tabcolsep}{6pt}
\renewcommand{\arraystretch}{0.9}
\begin{tabular}{@{}ll cc cc cc@{}}
\toprule
\multirow{2}{*}{\textbf{Encoder}} & \multirow{2}{*}{\textbf{Pre-training}} &
\multicolumn{2}{c}{\textbf{Stage (I vs Others)}} &
\multicolumn{2}{c}{\textbf{Vital Status}} &
\multicolumn{2}{c}{\textbf{Feature Quality (mean)}} \\
\cmidrule(lr){3-4} \cmidrule(lr){5-6} \cmidrule(lr){7-8}
& & {AUC $\uparrow$} & {Acc $\uparrow$} & {AUC $\uparrow$} & {Acc $\uparrow$} &
{\revblue{Feature diversity} (std) $\downarrow$} &
{Coverage $\uparrow$} \\
\midrule
\multirow{5}{*}{CONCH-V1~\cite{lu2024avisionlanguage}}
& MIL-NCE~\cite{miech2020end} & 0.692 & 0.719 & 0.754 & 0.857 & 0.999 ($3.0\text{e-}4$) & 0.065 \\
& LLM-loss~\cite{liu2023llava}& 0.724 & 0.688 & \textbf{0.759} & \textbf{0.882} & 0.841 (0.049) & \textbf{1.000} \\
& MoCoV3~\cite{chen2021empirical} & 0.705 & 0.594 & 0.554 & 0.863 & 0.999 ($4.1\text{e-}5$) & 0.288 \\
& Ours (no regularizer) & 0.741 & 0.672 & 0.714 & 0.845 & \textbf{0.110 (0.462)} & 0.812 \\
& \textbf{Ours (MAE)}& \textbf{0.763} & \textbf{0.734} & 0.696 & 0.845 & 0.346 (0.056) & 0.894 \\
\hline
Prov-GigaPath~\cite{xu2024gigapath} & \textit{MAE} (Slide-level)
& \revblue{0.728} & \revblue{0.688} & \revblue{0.709} & \revblue{0.845} & 0.382 (0.197) & {---} \\
\bottomrule
\end{tabular}

\label{tab:pretraining_resampler_comparison}
\end{table}

\subsection{Token Merging and Sparse Resampler Attention}
\label{app:stm_topk}

\myparagraph{Impact of STM.}
On the \emph{vision side}, using STM translates into large measured savings: with the same 60k‑token input, adding STM with $s{=}4$ reduces TFLOPs by about 14$\times$ and memory by about 3.7$\times$
 relative to the resampler‑only setting, without much performance decrease, shown in Tab.~\ref{tab:stm_ablation}. Also, the averaged score with $s{=}2$ slightly exceeds the no‑merge setting, with per‑task changes mostly within \mbox{$\pm$0.02}. Taken together with the large compute savings above, STM offers a favorable accuracy–efficiency trade‑off.
\begin{table}[!htbp]
  \caption{\textbf{STM impact on vision cost (TFLOPs, Memory) vs.\ final MLLM performance (WSI-Bench Avg).}}
\centering
{\footnotesize
\setlength{\tabcolsep}{6pt}
\renewcommand{\arraystretch}{0.95}
\begin{tabular}{@{}lcccc@{}}
\toprule
Setting & $s$ & TFLOPs $\downarrow$ & Mem (MB) $\downarrow$ & WSI-Bench Avg $\uparrow$ \\
\midrule
No Merge & 0 & 1.713 & 2254.5 & 0.735 \\
Merge 2x & 2 & 0.440 & 704.5 & \textbf{0.738} \\
Merge 4x & 4 & \textbf{0.121} & \textbf{607.1} & 0.737 \\
\bottomrule
\end{tabular}
}

\label{tab:stm_ablation}
\end{table}

\myparagraph{Impact of Top-$K$ Attention in the Resampler.}
We further study whether the sparse Top-$K$ cross-attention used in the resampler is necessary.
To this end, we replace the Top-$K$ operator with dense cross-attention while keeping all other components fixed (7B LLM, CONCH-V1 input tile-level tokens~\cite{lu2024avisionlanguage}, STM, $L{=}256$ latents and $M{=}96$ selected tokens).
As reported in Tab.~\ref{tab:detailed_ablation}, this dense resampler lowers the WSI-Bench average from $0.738$ of LoC-Path (7B) with Top-$K$ to $0.655$.
The degradation is consistent across all three task types.
These results confirm that enforcing sparsity via Top-$K$ attention is crucial for learning diverse, coverage-aware latents that support strong slide-level reasoning.

\Needspace*{16\baselineskip}  %
\subsection{Latent and Routing Budgets}
\label{app:budgets}
\nopagebreak\leavevmode\par
\begin{wrapfigure}{r}{0.27\textwidth}
\vspace{-1.0em}
\centering
\includegraphics[width=\linewidth]{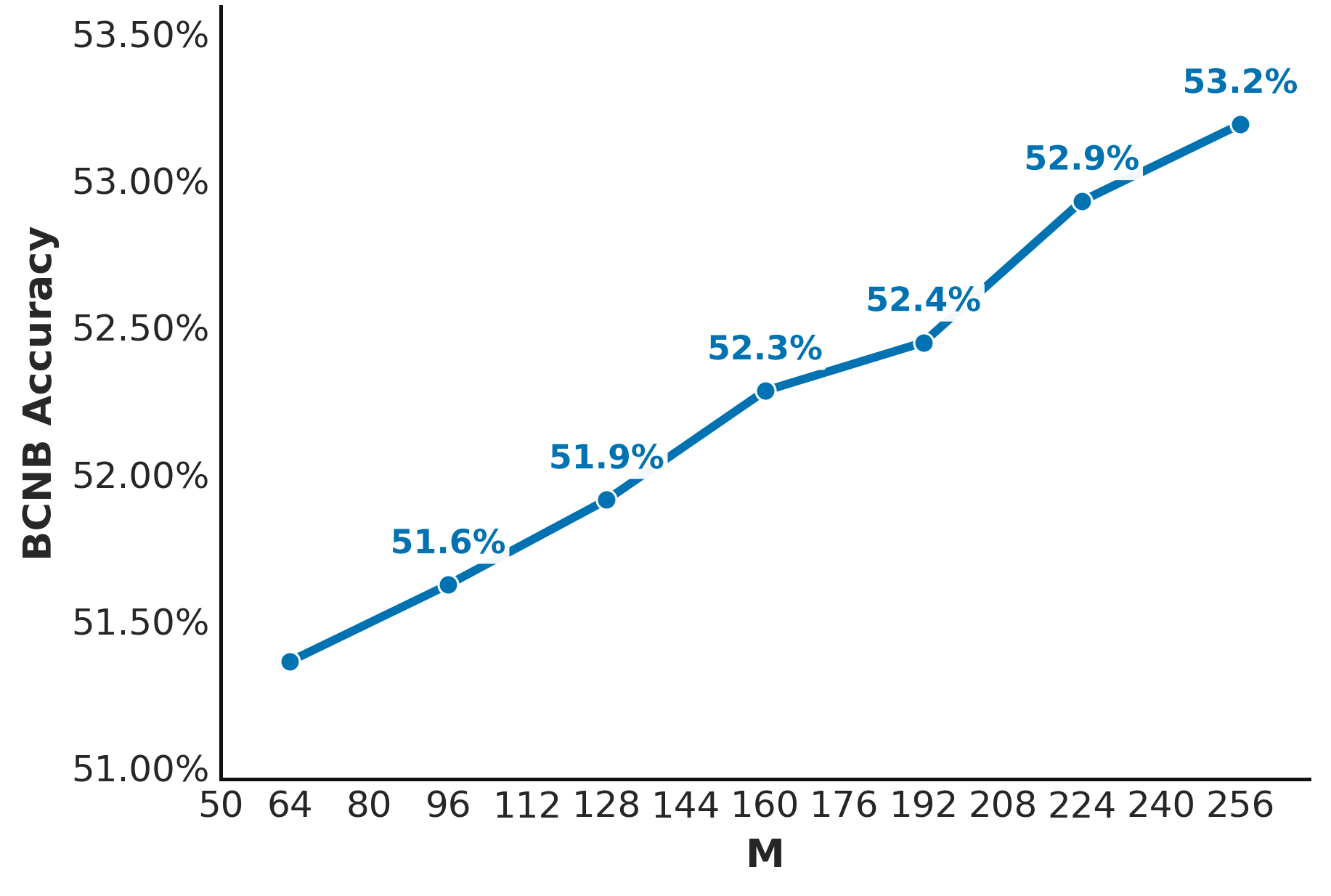}
\caption{BCNB accuracy trend as the selected token count $M$ increases.}
\label{fig:supp_bcnb_m}
\vspace{-1.0em}
\end{wrapfigure}

\nopagebreak\vspace{-\baselineskip}\nopagebreak  %
\begin{wraptable}[9]{r}{0.38\textwidth}  %
\vspace{-1.2em}
\centering
\footnotesize
\setlength{\tabcolsep}{6pt}
\caption{\textbf{Ablation on latent and routing budgets.} WSI-Bench average (excluding report generation) under different $(L,M)$ settings.}
\label{tab:ablation_compact_singlecol}
\begin{tabular}{lc}
\toprule
\(L\) / \(M\) & WSI-Bench Avg \\
\midrule
128 / 96      & 0.740 \\
256 / 96      & 0.738 \\
256 / 144     & 0.729 \\
\textbf{512} / 384  & \textbf{0.741} \\
\bottomrule
\end{tabular}
\vspace{-0.8em}
\end{wraptable}
\myparagraph{Impact of Latent and Routing Budgets.}
We vary the resampler’s latent length $L$ while keeping the Top-$M$ selected tokens active to capture language‑side complexity. On WSI‑Bench (except report generation), $L{=}512$ with $M{=}384$ attains the highest average (0.741), compared to others in Tab.~\ref{tab:ablation_compact_singlecol}. The trend indicates that increasing $L$ initially improves retrieval of rare but relevant regions for TIS, whereas overly large $L$ introduces redundancy and dilutes attention. If we keep $M$ fixed while increasing $L$, performance decreases slightly. On BCNB, increasing the selected token budget $M$ steadily improves accuracy (Fig.~\ref{fig:supp_bcnb_m}).

\myparagraph{WSI-LLaVA under Matched Token Budgets.}
Under our unified evaluation pipeline, we run the official WSI-LLaVA~\cite{liang2025wsi} checkpoint (Vicuna-7B LLM) and pool its visual tokens to smaller budgets (Tab.~\ref{tab:wsi_llava_token_pooling}). The reproduced full-budget VQA average ($0.746$) is slightly below the published $0.754$, and pooling to $L{=}256$ or $L{=}128$ changes it only marginally ($0.744$/$0.738$). At LoC-Path's matched budget ($L{=}128$), LoC-Path ($0.740$, Tab.~\ref{tab:main_results_wsi_bench}) is on par with WSI-LLaVA ($0.738$) while routing far fewer tokens to the LLM. This reinforces that the number of visual tokens is not the main quality lever in this regime.
\begin{table}[!b]
\centering
\caption{\textbf{WSI-LLaVA token-pooling results on WSI-Bench} (excluding report generation), reproduced under our unified evaluation pipeline. Metrics are open-ended WSI-Precision (P), WSI-Relevance (R), and closed-ended Accuracy (Acc) across morphological, diagnostic, and treatment-planning tasks.}
\label{tab:wsi_llava_token_pooling}
{\footnotesize
\setlength{\tabcolsep}{6pt}
\renewcommand{\arraystretch}{0.9}
\begin{tabular}{@{}l ccc ccc ccc c@{}}
\toprule
\multirow{2}{*}{\textbf{Model}} &
\multicolumn{3}{c}{\textbf{Morphological}} &
\multicolumn{3}{c}{\textbf{Diagnosis}} &
\multicolumn{3}{c}{\textbf{Treatment}} &
\multirow{2}{*}{\textbf{Avg.} $\uparrow$} \\
\cmidrule(lr){2-4}\cmidrule(lr){5-7}\cmidrule(lr){8-10}
& P & R & Acc & P & R & Acc & P & R & Acc & \\
\midrule
WSI-LLaVA ($L{=}576$) & 0.567 & \textbf{0.605} & \textbf{0.926} & \textbf{0.594} & \textbf{0.639} & 0.826 & \textbf{0.774} & \textbf{0.786} & \textbf{1.000} & \textbf{0.746} \\
WSI-LLaVA ($L{=}256$) & \textbf{0.568} & 0.604 & 0.924 & 0.588 & 0.632 & 0.843 & 0.759 & 0.780 & \textbf{1.000} & 0.744 \\
WSI-LLaVA ($L{=}128$) & 0.565 & 0.594 & 0.924 & 0.589 & 0.615 & \textbf{0.846} & 0.748 & 0.761 & \textbf{1.000} & 0.738 \\
\bottomrule
\end{tabular}

}
\end{table}

\subsection{Fusion Design: TIS and CARA}
\label{app:fusion_ablation}

\myparagraph{Impact of TIS Routing.}
We first isolate the value of query-dependent routing itself. In Tab.~\ref{tab:fusion}, adding TIS to a standard Flamingo-style cross-attention baseline improves the WSI-Bench average from $0.726$ to $0.729$, showing that routing is already helpful even with a simple fusion interface. In Tab.~\ref{tab:detailed_ablation}, removing TIS from our full CARA+TIS design reduces the average from $0.738$ to $0.736$. Although the drop is modest, the two comparisons are consistent: not all compressed resampler latents are equally useful for a given question, and TIS improves fusion by filtering the latent set before it reaches the LLM.

\myparagraph{TIS Under a Fixed Routing Budget.}
The ``Ours w/o TIS'' row of Tab.~\ref{tab:fusion} removes routing entirely and exposes all $L{=}256$ latents to CARA. To compare at the same budget, we additionally pool the 256 latents of the no-TIS model to $M{=}96$ by adaptive average pooling at inference. On the BCNB out-of-distribution test, TIS reaches an accuracy of 0.521 versus 0.496 for pooling (both models are re-decoded under a shared fixed-seed protocol for this paired test, which is why the TIS value differs slightly from the 0.519 in Tab.~\ref{tab:main_results_wsi_bench}), a paired WSI-cluster bootstrap difference of $+0.025$ (95\% CI $[0.017, 0.033]$, 10,000 replicates).\label{app:r2_tis}

\myparagraph{Impact of TIS Ranking Loss.}
Tab.~\ref{tab:detailed_ablation} isolates the effect of the ranking loss used to train TIS. Removing the pairwise ranking loss while keeping the distillation loss leads to a noticeable drop in WSI-Bench average from $0.738$ to $0.728$. The degradation is most visible on diagnosis and treatment planning: the closed-form diagnosis accuracy decreases from $0.860$ to $0.832$, and treatment planning also shows lower WSI-P/WSI-R. This suggests that $\mathcal{L}_{\mathrm{rank}}$ helps the TIS module learn \revblue{which latents to select, by aligning the relative TIS scores with the CARA teacher}, which improves cross-layer consistency.

\myparagraph{Impact of CARA Insertion Layers.}
We next study where to insert CARA inside the LLM. Tab.~\ref{tab:detailed_ablation} compares a single early CARA module (layer 1 only) with a distributed design using layers $\{1,8,15,22\}$. The distributed variant improves the WSI-Bench average from $0.730$ to $0.734$, indicating that injecting selected visual evidence at multiple depths is better than relying on a single early fusion point. Compared with our default early-to-mid placement $\{1,3,5,7\}$, however, the deeper distribution is still slightly weaker, suggesting that CARA is most effective in early-to-mid decoder layers rather than very deep layers.

\myparagraph{Impact of CARA Gating.}
We also study the learnable scalar gate $\gamma_\ell$ in CARA. In Tab.~\ref{tab:detailed_ablation}, fixing $\gamma_\ell=1$ at all layers yields a slightly lower WSI-Bench average ($0.736$) than the matched 7B main-paper baseline ($0.738$). The effect is modest but consistent with the role of gating: it allows each decoder layer to control how much selected visual evidence to absorb, instead of forcing full-strength fusion everywhere. We therefore keep the learnable gate in the default design.

\myparagraph{Sensitivity to $\lambda_{\mathrm{TIS}}$ and $\lambda_{\mathrm{rank}}$.}
Tab.~\ref{tab:r2_sens} gives the full results summarized in Sec.~\ref{sec:ablation}. Each row is a Stage-2 run of the $L{=}256$ model with one weight varied and the other fixed at its default ($\lambda_{\mathrm{TIS}}{=}0.02$, $\lambda_{\mathrm{rank}}{=}0.5$), evaluated on the WSI-Bench closed-task test sets, BCNB, and PathBench.\label{app:r2_sens}
\begin{table}[!htbp]\centering\footnotesize
\caption{\revblue{Sensitivity to the fusion-loss weights (accuracy). Top: $\lambda_{\mathrm{TIS}}$ with $\lambda_{\mathrm{rank}}{=}0.5$; bottom: $\lambda_{\mathrm{rank}}$ with $\lambda_{\mathrm{TIS}}{=}0.02$.}}
\label{tab:r2_sens}
\begin{tabular}{@{}lcccccc@{}}\toprule Setting & Morph. & Diag. & Treat. & Closed macro & BCNB & PathBench \\\midrule
$\lambda_{\mathrm{TIS}}{=}0.01$ & 0.913 & 0.754 & 0.917 & 0.861 & 0.490 & 0.797 \\
$\lambda_{\mathrm{TIS}}{=}0.02$ & 0.943 & 0.860 & 0.979 & 0.927 & 0.519 & 0.813 \\
$\lambda_{\mathrm{TIS}}{=}0.05$ & 0.922 & 0.764 & 0.875 & 0.854 & 0.476 & 0.786 \\
\midrule
$\lambda_{\mathrm{rank}}{=}0$ & 0.934 & 0.832 & 0.979 & 0.915 & 0.507 & 0.797 \\
$\lambda_{\mathrm{rank}}{=}0.05$ & 0.933 & 0.857 & 0.979 & 0.923 & 0.515 & 0.810 \\
$\lambda_{\mathrm{rank}}{=}0.5$ & 0.943 & 0.860 & 0.979 & 0.927 & 0.519 & 0.813 \\
$\lambda_{\mathrm{rank}}{=}5.0$ & 0.933 & 0.849 & 0.979 & 0.920 & 0.502 & 0.823 \\
\bottomrule\end{tabular}\end{table}

\subsection{Tile-level Encoders and LLM Scale}
\label{app:backbones}

\myparagraph{Impact of Tile-level Encoders.}
We primarily use CONCH-V1~\cite{lu2024avisionlanguage} 512-dim features as tile-level inputs, but our framework is compatible with other tile-level vision encoders.
Tab.~\ref{tab:pretraining_resampler_comparison} already shows that the MAE-pretrained resampler produces high-quality latents from CONCH-V1 features~\cite{lu2024avisionlanguage}, remaining competitive with the slide-level encoder upper bound on classification tasks.
To further test robustness, we swap the tile encoder to CONCH-V15~\cite{ding2025titan}, whose feature dimension is $768$.
To accommodate this higher dimension, we increase the resampler hidden size to $768$ and use three cross-attention layers.
As summarized in Tab.~\ref{tab:detailed_ablation}, the CONCH-V15~\cite{ding2025titan} variant improves the overall WSI-Bench average to $0.745$, especially boosting WSI-Relevance for diagnosis and treatment planning.
This suggests that LoC-Path can directly benefit from advances in pathology foundation models without changing the overall MLLM architecture.
We also swap the tile-level encoder to a DINOv2~\cite{oquab2023dinov2} variant without feature alignment pre-training, which yields worse overall performance on WSI-Bench tasks. This suggests that the feature space alignment training is essential for bridging the vision encoders and the pre-trained LLMs.  

\myparagraph{Impact of LLM Scale.}
Finally, we scale the language backbone from Qwen2.5-7B-Instruct (used in the main paper) to Qwen2.5-14B while keeping the visual stack and routing modules fixed~\cite{qwen25}.
Keeping the visual stack fixed at $L{=}256$ and $M{=}96$, scaling the language backbone from Qwen2.5-7B-Instruct to Qwen2.5-14B improves the WSI-Bench average from $0.738$ to $0.744$.
The improvement is driven mainly by treatment planning, while morphological analysis and diagnosis change only slightly, indicating that our token-compression pipeline already supplies a strong visual representation even for a relatively small LLM.
Scaling up the language model yields additional accuracy, but it is less noticeable compared to the gain by switching to a better tile-level encoder.

\subsection{Pipeline-level Lesion Preservation on CAMELYON16}
\label{app:r2_lesion}

This paragraph details the small-lesion analysis in Sec.~\ref{sec:ablation}. We use the official CAMELYON16 split (270 training and 129 test slides) and the official mask-based FROC protocol, without fine-tuning any LoC-Path module. At each compression stage, a linear lesion probe is trained on the training split: on raw CONCH tile features, on STM-merged group features (73.8\% fewer tokens than the raw tiles), and on a spatial back-projection of the resampler latents. Probes are trained with three seeds, with the regularization strength selected on held-out slides; confidence intervals are from 10,000 whole-slide paired bootstrap replicates. The small-lesion subset consists of the 31 tumor test slides whose annotated lesions cover less than 2\% of tissue tiles, evaluated together with the 80 normal test slides.
\begin{table}[!htbp]\centering\footnotesize
\caption{\revblue{Official-mask FROC of linear lesion probes at each compression stage (mean $\pm$ SD over three seeds; 95\% whole-slide bootstrap CI) and paired differences.}}
\label{tab:r2_froc}
\begin{tabular}{@{}llcc@{}}\toprule Subset & Stage & FROC & 95\% CI \\\midrule
All 129 test slides & Raw CONCH tiles & 0.548 $\pm$ 0.044 & $[0.447, 0.657]$ \\
 & STM groups & 0.676 $\pm$ 0.028 & $[0.576, 0.779]$ \\
 & Resampler back-projection & 0.268 $\pm$ 0.043 & $[0.153, 0.397]$ \\
 & $\Delta$ STM $-$ raw & $+0.126$ & $[0.058, 0.185]$ \\
 & $\Delta$ resampler $-$ STM & $-0.409$ & $[-0.523, -0.304]$ \\
\midrule
Small-lesion subset (31 tumor + 80 normal slides) & Raw CONCH tiles & 0.624 $\pm$ 0.050 & $[0.507, 0.744]$ \\
 & STM groups & 0.744 $\pm$ 0.022 & $[0.627, 0.845]$ \\
 & Resampler back-projection & 0.065 $\pm$ 0.009 & $[0.010, 0.133]$ \\
 & $\Delta$ STM $-$ raw & $+0.118$ & $[0.046, 0.187]$ \\
 & $\Delta$ resampler $-$ STM & $-0.677$ & $[-0.772, -0.574]$ \\
\bottomrule\end{tabular}\end{table}

Tab.~\ref{tab:r2_froc} shows that STM preserves the lesion signal: although the fixed $2{\times}2$ merge mixes lesion and non-lesion tiles, the probe FROC improves over raw tiles both overall (0.548 to 0.676) and on the small-lesion subset (0.624 to 0.744) with 73.8\% fewer tokens. After resampling, 99.78\% of the lesion-positive groups on the small-lesion slides receive attention from at least one latent, and lesion groups receive 2.77$\times$ their proportional attention mass; mapping the $M{=}96$ latents routed by TIS back to tiles through the inverse STM mapping shows that 96.90\% of lesion tiles on all tumor slides and 98.75\% on the small-lesion slides remain covered. The much lower FROC of the back-projection probe indicates that lesion location is not recoverable from the global latents by this simple probe. Compression is therefore lossy at the localization level; what the MAE-trained resampler preserves is that lesion tiles remain within its attention support, and we claim retention only in this sense.

\FloatBarrier
\section{Interpretation and Qualitative Results}
\label{app:more_qualitative_results}

\subsection{Construction of Fig.~\ref{fig:tis_attention}: Token-Mass Maps and Weak Tissue Labels}
\label{app:tissue_labeling}

\begin{table}[!htbp]
\centering
\small
\setlength{\tabcolsep}{4pt}
\renewcommand{\arraystretch}{1.08}
\caption{\textbf{Thirteen coarse tissue classes used for weak semantic labeling in Fig.~\ref{fig:tis_attention}.}
The taxonomy and operational definitions were specified with guidance from a practicing pathologist.
These categories are used only for interpretability analysis and do not constitute definitive pathology labels.}
\label{tab:tissue_classes}
\begin{tabularx}{\linewidth}{@{}lX@{}}
\toprule
\textbf{Class} & \textbf{Operational definition} \\
\midrule
malignant tissue & Regions showing morphology consistent with malignant neoplasia, such as marked atypia and/or infiltrative growth patterns. \\
benign tissue & Non-malignant tissue regions without evident malignant infiltrative morphology. \\
stroma & Stromal background components, including fibrous, desmoplastic, or other supporting connective tissue. \\
lymphocytes & Regions predominantly characterized by lymphocytic infiltration. \\
necrosis & Necrotic tissue regions with loss of viable architecture or cellular detail. \\
adipose tissue & Regions dominated by adipocytes (fat tissue). \\
tissue artifact & Slide, preparation, or scanning artifact regions, such as folds, tears, blur, or staining artifacts. \\
blood vessel & Vascular regions with identifiable vessel-related structures, such as lumen or vessel-wall patterns. \\
extracellular mucin & Regions dominated by extracellular mucin pools. \\
nerve & Nerve tissue or nerve bundle regions. \\
hemorrhage & Regions dominated by hemorrhage or extravasated blood. \\
smooth muscle & Regions dominated by smooth muscle components. \\
plasma cells & Regions enriched with plasma cells. \\
\bottomrule
\end{tabularx}
\end{table}

Fig.~\ref{fig:tis_attention} is a representative \emph{post-hoc} case study used to visualize how TIS redistributes evidence after routing. It does not affect model training or inference. This section explains how the heatmaps and tissue-level bar charts in Fig.~\ref{fig:tis_attention}(b--e) are constructed.

\myparagraph{Patch-level Token Mass.}
Let $p_{q,n}$ denote the resampler cross-attention probability from latent $q$ to patch/token $n$ after averaging over attention heads, where $q=1,\ldots,Q$ and $n=1,\ldots,N$.
We define the \emph{Post-Resampler} patch mass as the average over all resampler latents:
\[
m_n^{\mathrm{res}}
=
\frac{1}{Q}\sum_{q=1}^{Q} p_{q,n}.
\]
For \emph{Post-TIS}, let $\mathcal I=(i_1,\ldots,i_{M_{\mathrm{plot}}})$ be the ordered selected latent indices used for visualization, and let $\rho_{i_k}$ be their TIS scores. In Fig.~\ref{fig:tis_attention}, we use $M_{\mathrm{plot}}=16$ for visualization, while the training-time routing budget remains $M=96$. We first normalize the TIS scores,
\[
w_k=\operatorname{softmax}_k(\rho_{i_k}),
\]
and then form the Post-TIS patch mass by a weighted average over the selected latents:
\[
m_n^{\mathrm{tis}}
=
\sum_{k=1}^{M_{\mathrm{plot}}} w_k\, p_{i_k,n}.
\]
Both $m^{\mathrm{res}}$ and $m^{\mathrm{tis}}$ sum to one over patches/tokens. Panels (b) and (c) visualize these two patch-mass maps using a shared within-case color scale. This display normalization is only for visualization and does not affect the bar plots.

\myparagraph{Weak Semantic Tissue Labels.}
To obtain the tissue-level summaries in Fig.~\ref{fig:tis_attention}(d,e), we assign each patch/token a weak semantic tissue label from the fixed vocabulary of 13 coarse classes listed in Tab.~\ref{tab:tissue_classes}. These labels are intended only for interpretability analysis and do not constitute definitive pathology diagnoses.

We obtain the labels in a zero-shot manner using the CONCH text encoder~\cite{lu2024avisionlanguage}. For each tissue class $c$, we construct a fixed bank of 13 text templates by replacing \texttt{CLASSNAME} with the class name, giving $13\times 13=169$ prompts in total across all classes. Let $\mathcal T$ denote the template set. We encode and average the prompts to form a class prototype:
\[
\bm e_c
=
\operatorname{Normalize}\!\left(
\frac{1}{|\mathcal T|}
\sum_{t\in\mathcal T}
\operatorname{Normalize}(\mathrm{TextEnc}(\texttt{prompt}_{c,t}))
\right).
\]
For each normalized patch/token feature $\hat{\bm v}_n$, we compute cosine similarities
\[
s_{n,c} = \cos(\hat{\bm v}_n,\bm e_c),
\]
and assign the weak semantic label
\[
y_n = \arg\max_c s_{n,c}.
\]
We then aggregate the patch masses within each predicted class:
\[
M_c^{\mathrm{res}}=\sum_{n:\,y_n=c} m_n^{\mathrm{res}},
\qquad
M_c^{\mathrm{tis}}=\sum_{n:\,y_n=c} m_n^{\mathrm{tis}}.
\]
These class-wise masses are plotted in Fig.~\ref{fig:tis_attention}(d).

\myparagraph{Report-derived Relevant Tissue Classes.}
For Fig.~\ref{fig:tis_attention}(e), we further define a report-derived relevant class set $\mathcal R$ from the reference report text.
Specifically, we use a simple keyword-matching heuristic to map the report to one or more tissue classes.
Examples include \emph{carcinoma/tumor/cancer} for \emph{malignant tissue}, \emph{necrosis/necrotic} for \emph{necrosis}, \emph{vascular/venous/arterial/lymphovascular} for \emph{blood vessel}, and \emph{adipose/fat} for \emph{adipose tissue}.
If no keyword is matched, we default to \{\emph{malignant tissue}\}.
The bars in Fig.~\ref{fig:tis_attention}(e) show the class-wise masses for $c\in\mathcal R$, and the reported \emph{total relevant mass} is
\[
M_{\mathcal R}^{\mathrm{res}}=\sum_{c\in\mathcal R} M_c^{\mathrm{res}},
\qquad
M_{\mathcal R}^{\mathrm{tis}}=\sum_{c\in\mathcal R} M_c^{\mathrm{tis}}.
\]

\noindent\textbf{Note.}
Both the tissue labels and the report-derived relevant class set are weak, heuristic constructs used only for post-hoc interpretability.
Accordingly, Fig.~\ref{fig:tis_attention}(e) should be interpreted as a weak report-alignment analysis rather than a ground-truth semantic attribution result.

\FloatBarrier

\subsection{Latent Diversity and Coverage}
\label{app:qual_latent}

\begin{figure}[!htbp]
  \centering
  \includegraphics[width=\linewidth]{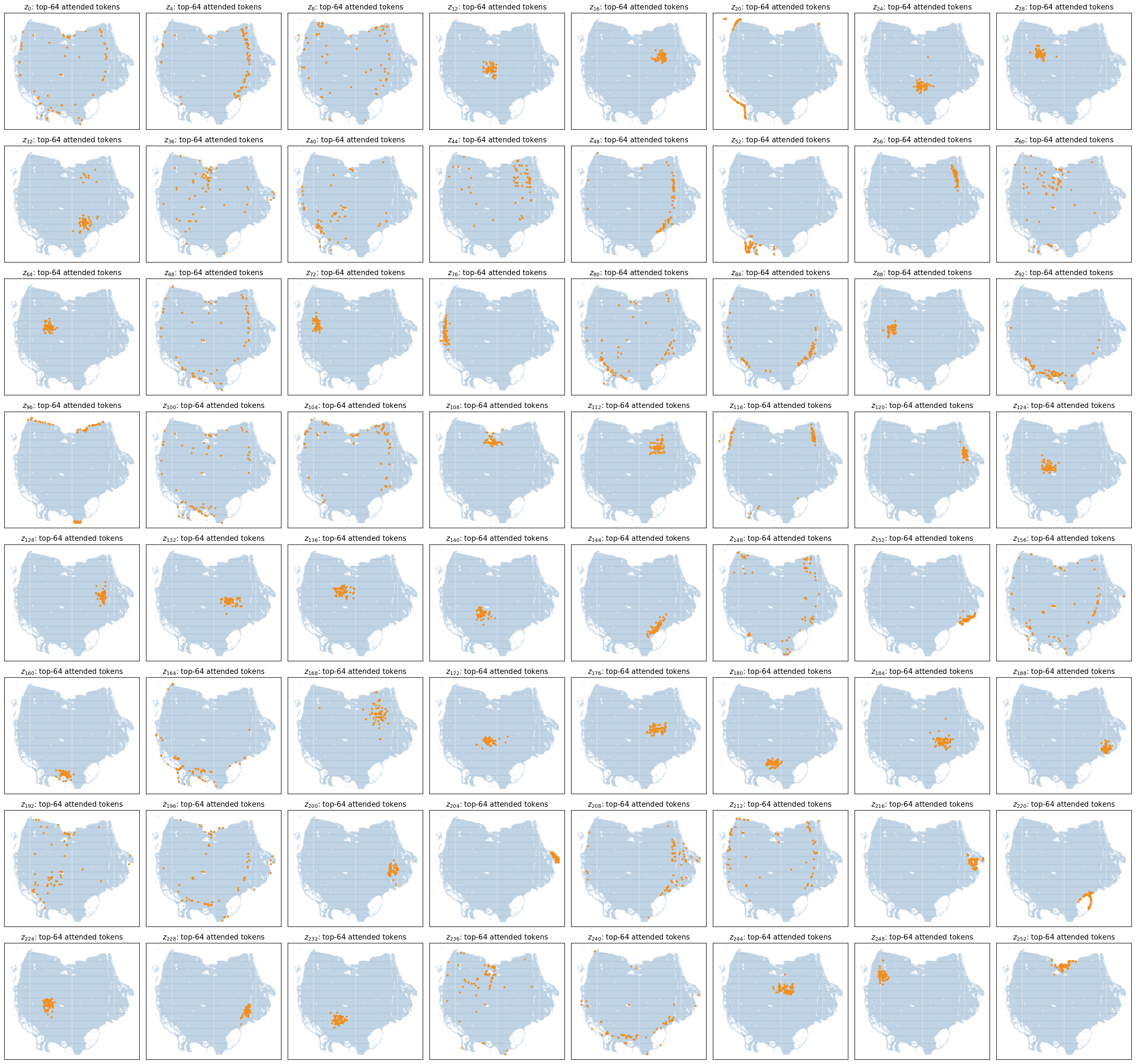}
  \caption{\textbf{Diversity of resampler latents via Top-$K$ attention.} 
  Each panel corresponds to one latent from the MAE-pretrained resampler. This resampler has 256 output latents. 
  Orange dots mark the spatial locations of the Top-64 input tokens receiving the highest
  attention from that latent; the blue background shows all foreground token positions.
  Different latents specialize to distinct regions and structures while jointly covering the
  whole slide, qualitatively confirming the high coverage and diversity reported in
  Sec.~\ref{sec:results} and Tab.~\ref{tab:pretraining_resampler_comparison}.}
  \label{fig:latent_diversity}
\end{figure}

To further analyze the behavior of the resampler latents, Fig.~\ref{fig:latent_diversity} visualizes Top-$K$ attention maps for multiple latents on the same WSI, complementing Fig.~\ref{fig:covery_and_attention_map}.
In Fig.~\ref{fig:latent_diversity}, different latents focus on distinct tissue regions and morphological patterns.
At the same time, they cover most of the slide, matching the high coverage reported in Sec.~\ref{sec:results}.
Fig.~\ref{fig:latent_diversity} confirms that the resampler indeed “glances” over the whole WSI and allocates
different latents to complementary visual contexts rather than redundantly attending to the same locations.

\subsection{MAE Resampler Reconstruction}
\label{app:qual_mae}

\begin{figure}[!htbp]
  \centering
  \includegraphics[width=0.8\linewidth]{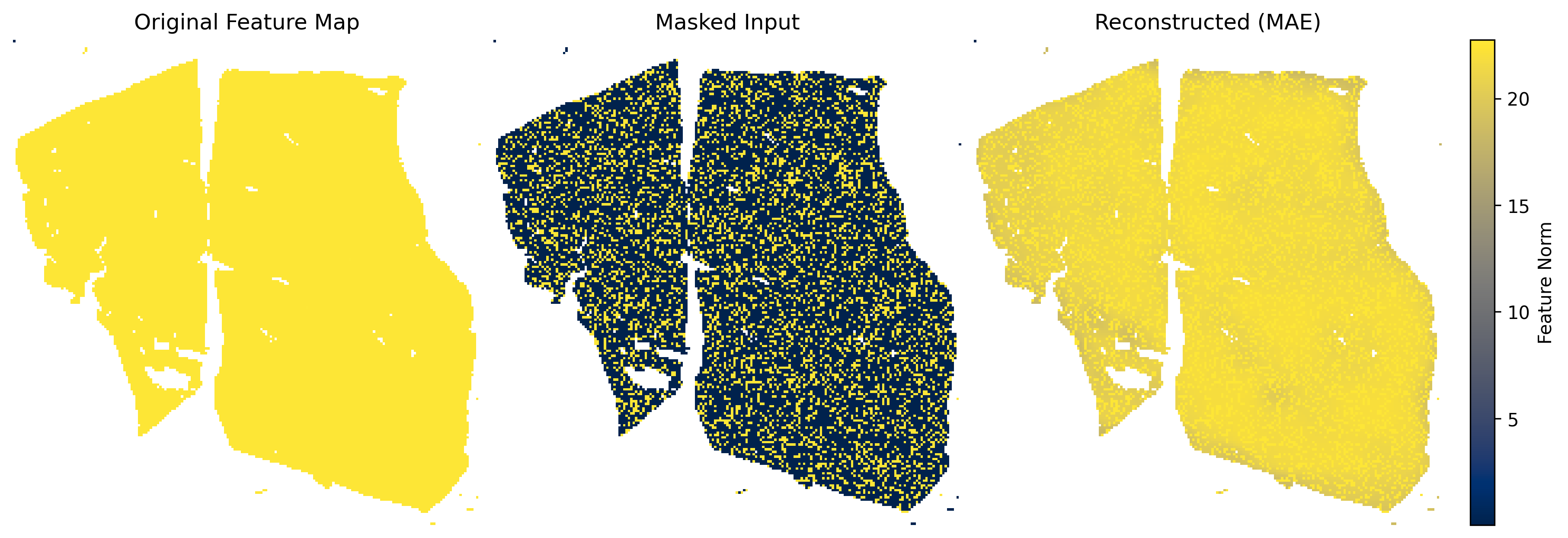}
  \caption{\textbf{Visualization of MAE resampler reconstruction.}
  Left: Original tile-level feature map (visualized by feature norm). Middle: Masked input (\revblue{70\%} mask ratio); black regions indicate missing tokens. Right: Features reconstructed by the MAE-pretrained resampler (256 latents). Despite heavy corruption, the model successfully recovers the global tissue layout and feature magnitude, demonstrating the efficacy of the compressed latent representation.}
  \label{fig:mae_recon}
\end{figure}

To better illustrate what the MAE-pretrained resampler learns, Fig.~\ref{fig:mae_recon}
visualizes an example WSI before and after reconstruction.
The left panel shows the original tile-level feature map, with feature norms encoded as a heatmap
over the 2D grid.
The middle panel shows the masked input used during pretraining, where \revblue{$70\%$} of the tiles are
randomly removed.
The right panel presents the reconstruction from our resampler decoder.
Despite the heavy corruption, the model successfully recovers the global tissue layout and the
overall magnitude of tile features, supporting our assumption that a small set of latents can
faithfully encode the slide-level representation.
This qualitative view complements the coverage visualization in Fig.~\ref{fig:covery_and_attention_map} and the latent-quality statistics in Tab.~\ref{tab:pretraining_resampler_comparison}.

\subsection{STM Merged Grid Visualization}
\label{app:qual_stm}

\begin{figure}[!htbp]
  \centering
  \includegraphics[width=0.8\linewidth]{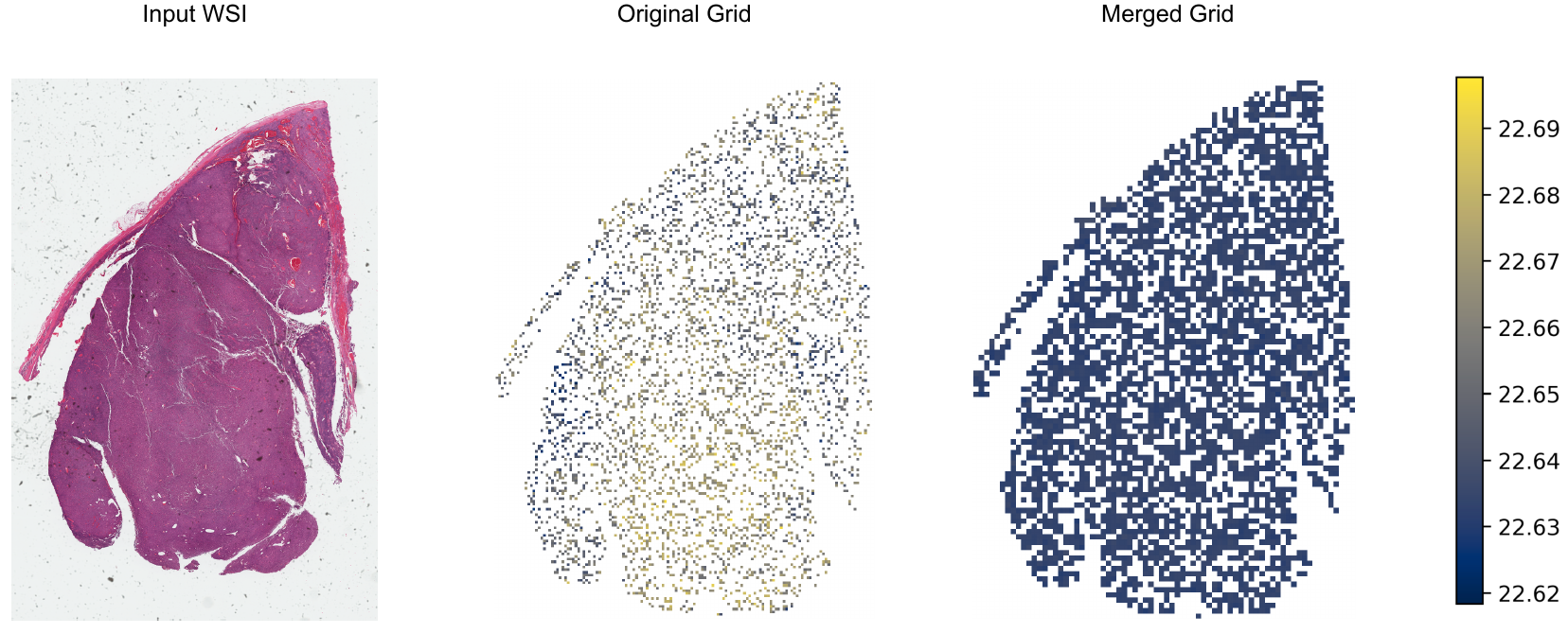}
  \caption{\textbf{STM merging behavior on a TCGA WSI.}
  From left to right: (i) input WSI thumbnail, (ii) original tile grid with per-tile feature norms,
  and (iii) merged grid after applying STM with a local window (e.g., $s{=}2$).
  STM aggressively reduces the number of tiles while largely preserving the overall tissue footprint
  and coarse morphology, illustrating how local redundancy can be removed before resampling.}
  \label{fig:stm_merge}
\end{figure}

Fig.~\ref{fig:stm_merge} provides a visual explanation of how STM
removes local redundancy.
In Fig.~\ref{fig:stm_merge}, the right panel shows the merged grid after applying STM with a local window ($s{=}2$).
Although the number of tokens is reduced by roughly $4\times$, the merged tokens still form a dense
coverage of the tissue region and preserve coarse anatomical structures.
This visualization is consistent with the quantitative results in Tab.~\ref{tab:detailed_ablation} and Tab.~\ref{tab:stm_ablation}, where STM does not hurt accuracy while dramatically reducing FLOPs and memory on the vision side.

\subsection{Task-level Successes and Failure Modes}
\label{app:qual_fail}

\begin{figure}[!htbp]
    \centering
    \includegraphics[width=0.8\linewidth]{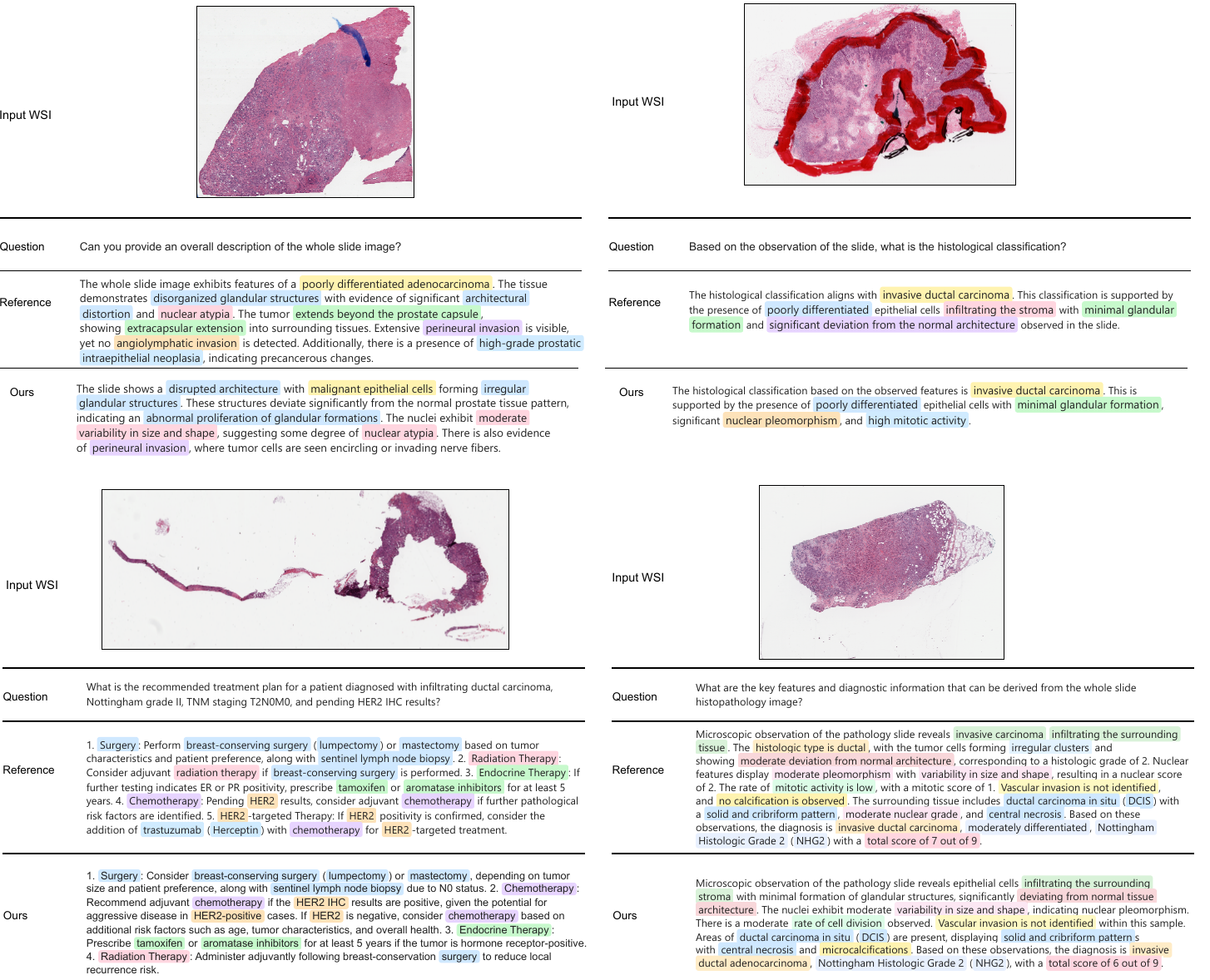}
    \caption{\textbf{Qualitative success and failure cases across WSI-Bench tasks.} 
    Top row: Successful cases in morphological analysis and diagnosis, showing high overlap with expert descriptions. 
    Bottom row: Typical failure modes in treatment planning and report generation. While the model captures global intent, it may miss targeted options or hallucinate details.
    (Highlight colors mark corresponding content in the reference and our output.)}
    \label{fig:qual_success_failure}
\end{figure}

Fig.~\ref{fig:qual_success_failure} illustrates LoC-Path's performance across four tasks in WSI-Bench. Regarding successes (top row), the model accurately identifies dominant tumor types and key morphological patterns, such as invasive ductal carcinoma, infiltrative growth, and nuclear pleomorphism. This alignment with expert reports confirms that our compression preserves clinically salient regions. However, failure modes exist (bottom row). In treatment planning, the model tends to offer generic guideline-based regimens while under-emphasizing case-specific constraints like biomarkers. Similarly, in report generation, while correctly identifying tumor type and grade, it occasionally hallucinates fine-grained attributes (e.g., Nottingham scores or microcalcifications). Overall, LoC-Path reliably captures coarse diagnostic intent but lacks precision for granular decision-critical details, which is also consistent with the earlier quantitative results.

\FloatBarrier
\section{Judge Prompts for WSI-Precision and WSI-Relevance}
\label{app:judge_prompts}

\myparagraph{Claims Extraction Prompt.}

\begin{Verbatim}[fontsize=\scriptsize,frame=single,framesep=2mm]
System Message: You are an AI assistant specialized in
processing pathological diagnosis Q&A pairs.
I will provide you with a pathology diagnosis question and
its corresponding answer.

Your task is to:

Claims Extraction:
- Carefully analyze the answer and remove any unnecessary
  information that is not directly relevant to the question.
- Only extract claims that directly address the question.
  Discard any information that does not directly answer or
  pertain to the question.
- Divide the refined answer into several distinct and
  granular claims.
- Keep closely related information together in the same
  claim to preserve context and meaning.
  Do not split sentences or ideas that are logically
  connected.
- Break down complex sentences into smaller, individual
  claims only if it does not disrupt the logical flow or
  separate connected ideas.
- Ensure there is no omission or repetition among the
  claims.

Guidelines:
- Only output the claims without including any additional
  text or explanations.
- Each claim should be concise and represent a single fact
  or point directly related to the question.
- Maintain the integrity of statements that are
  contextually connected.

Output Format: Present the extracted claims as a list in the
following format:
["claim1", "claim2", "claim3", ...]
\end{Verbatim}

\noindent\begin{minipage}[t]{0.49\textwidth}
\myparagraph{WSI-Precision Prompt.}

\begin{Verbatim}[fontsize=\fontsize{6.5}{7.8}\selectfont,frame=single,framesep=2mm]
System Message: Please act as an impartial judge and
evaluate the correctness of the AI assistant's pathology
dialogue for each claim based on the following scoring
criteria. Provide an explanation for each evaluation and
assign a score.

**Scoring Criteria:**
- **1**: The information in the pathology dialogue is
  completely correct regarding the claim.
- **0.7**: The information is mostly correct and closely
  aligns with the claim.
- **0.3**: The claim is mentioned but contains errors in the
  core content (e.g., mistakes in differentiation degree or
  malignancy).
- **0**: The information in the pathology dialogue is
  completely incorrect regarding the claim.

Output Requirements:

Please output your evaluations as a list of dictionaries in
plain text format (not JSON). The format should be as
follows:

[
  {
    "claim": "Original claim1",
    "explanation": "Explanation for the score",
    "score": 1 or 0.7 or 0.3 or 0
  },
  {
    "claim": "Original claim2",
    "explanation": "Explanation for the score",
    "score": 1 or 0.7 or 0.3 or 0
  },
  ...
]
\end{Verbatim}
\end{minipage}\hfill
\begin{minipage}[t]{0.49\textwidth}
\myparagraph{WSI-Relevance Prompt.}

\begin{Verbatim}[fontsize=\fontsize{6.5}{7.8}\selectfont,frame=single,framesep=2mm]
System Message: Please act as an impartial judge and
evaluate the relevance of the original ground truth answer
to each claim derived from the model's answer.
Provide an explanation for each evaluation and assign a
score based on the following criteria.

**Scoring Criteria:**
- **1**: The content in the ground truth answer is
  completely relevant to the claim.
- **0.7**: The content is mostly relevant but has minor
  omissions or deviations.
- **0.3**: The content is partially relevant with
  significant omissions or irrelevant information.
- **0**: The content in the ground truth answer is not
  relevant to the claim.

Output Requirements:

Please output your evaluations as a list of dictionaries in
plain text format (not JSON). The format should be as
follows:
[
  {
    "claim": "Original claim1",
    "explanation": "Explanation for the score",
    "score": 1 or 0.7 or 0.3 or 0
  },
  {
    "claim": "Original claim2",
    "explanation": "Explanation for the score",
    "score": 1 or 0.7 or 0.3 or 0
  },
  ...
]
\end{Verbatim}
\end{minipage}